\documentclass[journal]{IEEEtran}

\usepackage{dblfloatfix}			
\usepackage{enumitem} 				

\usepackage{cite}
\usepackage{url}

\usepackage{soul}
\usepackage{graphicx}
\usepackage{graphbox}
\usepackage{capt-of}
\usepackage{epsfig} 				
\graphicspath{{figure/}}		 	
\usepackage[export]{adjustbox} 	 	
\usepackage[caption=false,
			font=normalsize,
			labelfont=sf,
			textfont=sf]{subfig}	
\usepackage{mathtools}

\usepackage{amsmath}	
\usepackage{amssymb}	

\usepackage{algorithm}

\usepackage[table,xcdraw]{xcolor}
\usepackage{tabularx}
\usepackage{booktabs}
\usepackage{multirow}			 	
\usepackage{array}					
\newcolumntype{L}[1]{>{\raggedright\let\newline\\\arraybackslash\hspace{0pt}}m{#1}}
\newcolumntype{C}[1]{>{\centering\let\newline\\\arraybackslash\hspace{0pt}}m{#1}}
\newcolumntype{R}[1]{>{\raggedleft\let\newline\\\arraybackslash\hspace{0pt}}m{#1}}
\renewcommand{\arraystretch}{1.3}

\usepackage{etoolbox}

\newlength\Origarrayrulewidth
\newcommand\Thickvrule[1]{%
\multicolumn{1}{!{\color{red}\vrule width 1pt}c!{\color{red}\vrule width 1pt}}{#1}%
}

\begin{document}

\title{CDN-MEDAL: Two-stage Density and Difference Approximation Framework for Motion Analysis}


\author{\large
	Synh Viet-Uyen Ha$^{\ast}$,~\IEEEmembership{Member,~IEEE,}
	Cuong Tien Nguyen$^{\ast}$,
	Hung Ngoc Phan$^{\ast}$,
	Nhat Minh Chung,\\
	and~Phuong Hoai Ha~\IEEEmembership{Member,~IEEE,}

	\thanks{S. V.-U. Ha, C. T. Nguyen, H. N. Phan and N. M. Chung are with the School of Computer Science and Engineering, International University, Vietnam National University, Ho Chi Minh City, Vietnam.}
	\thanks{P. H. Ha is with the Department of Computer Science, UiT The Arctic University of Norway.}
	\thanks{\noindent $^{\ast}$Equal distributions with corresponding email: hvusynh@hcmiu.edu.vn}
}

\maketitle

\begin{abstract}
Background modeling and subtraction is a promising research area with a variety of applications for video surveillance. Recent years have witnessed a proliferation of effective learning-based deep neural networks in this area. However, the techniques have only provided limited descriptions of scenes' properties while requiring heavy computations, as their single-valued mapping functions are learned to approximate the temporal conditional averages of observed target backgrounds and foregrounds. On the other hand, statistical learning in imagery domains has been a prevalent approach with high adaptation to dynamic context transformation, notably using Gaussian Mixture Models (GMM) with its generalization capabilities. By leveraging both, we propose a novel method called CDN-MEDAL-net for background modeling and subtraction with two convolutional neural networks. The first architecture, CDN-GM, is grounded on an unsupervised GMM statistical learning strategy to describe observed scenes' salient features. The second one, MEDAL-net, implements a light-weighted pipeline of online video background subtraction. Our two-stage architecture is small, but it is very effective with rapid convergence to representations of intricate motion patterns. Our experiments show that the proposed approach is not only capable of effectively extracting regions of moving objects in unseen cases, but it is also very efficient.
\end{abstract}


\IEEEpeerreviewmaketitle

\vspace{-2mm}
\section{Introduction}
With the swift progress in computer vision, practical utilization of static cameras in surveillance systems are promising technologies for advanced tasks such as behavior analysis \cite{Zhang2020}, object segmentation \cite{Ammar2020} and motion analysis \cite{Park2018}, \cite{Bilge2017}. Among the various functionalities, online background modeling and subtraction is a pivotal component for proper understanding of scene dynamics and extraction of attributes of interest. An ideal background is a scene containing only stationary objects and features uninteresting to the system (e.g., streets, houses). Then, by comparing visual inputs from a video sequence with their backgrounds, desired objects, called foregrounds (e.g., cars, pedestrians), can be localized for analysis. As real-life scenarios involve various degrees of dynamics like illumination changes or scene motions, there are many approaches for constructing backgrounds and detecting changes \cite{Bouwmans2014}.

One prominent method to handle the background modeling and subtraction problems with statically mounted cameras is to employ pixel-based statistical frameworks such as Gaussian Mixture Models (GMM) \cite{Stauffer1999}, \cite{Zivkovic2004}, \cite{Martins2018}, \cite{Ha2020}. These GMM frameworks are based on the hypothesis that background intensities are observed most frequently in a video sequence recorded from a still camera, thereby creating explicit mathematical structures for exploitation. Additionally, notable characteristics of such designs include their applicability even under illumination changes (e.g. from moving clouds), view noises (e.g. rain, snow flakes), and implicit motions (e.g. river water). However, these methods struggle in generalizing when the background intensity hypothesis fails (e.g. stopped objects), resulting in corruptive backgrounds, which often leads to inaccurate foreground estimations. Furthermore, these statistical techniques tends to involve a sequential computing paradigm, and lack explicit parallelism for dealing with big data.

On the other hand, ever-advancing processing units specialized for large-scale data is making Deep Neural Networks (DNNs) a prominent pattern extraction and visual prediction mechanism. Through data-driven learning, deep learning approaches for motion detection in videos are rapidly showcasing not only their effectiveness, but also their efficient utilizations of parallel computing technologies. However, architectures for background modeling and foreground detection are computationally expensive as a trade-off for high accuracies. Furthermore, DNNs' two critical shortfalls have been remarked:

\textit{A requirement of a huge-scale dataset of labeled images}: DNN-based models for motion detection exploit weak statistical regularities between input images and labelled background scenes. In this case, a prohibitively large dataset with all practical scenarios and their labels is required for an all-scenario generalization. With the lack of background labels for training \cite{Kalsotra2019}, there is currently no universal data-driven experiment to assure that the scenes' true properties are appropriately presented.

\textit{A prevailing fail on contextual variation}: Recently, binary classification schemes with DNNs have been applied to online foreground segmentation. Minimizing the sum-of-squares or cross-entropy error function has been proposed to demonstrate the performance of DNNs-based approaches for foreground extraction in various scenarios. In this approach, models are trained to represent the semantic properties on the training sets when the target of generalization is a specific experimental video sequence. For various unseen contextual dynamics that may occur in the real world \cite{Bouwmans2019, Geoffrey2015}, this conditional average will be inadequate. Essentially, DNN-based methods excel in the experimental datasets of background modelling and change detection but may fail in real-world scene variations.

Nevertheless, DNN-based approaches are very promising
as they can approximate any functions up to any arbitrary accuracies. This signifies that we can utilize their parallelism to not only approximate the mechanism behind the optimization of GMM for background modeling, but also facilitate an efficient and complex data-driven foreground extractor. 
In this article, we circumvent DNNs' issues while reserving their benefits by including explicit modelling of GMM into our novel, light-weighted, dual framework of two convolutional neural networks (CNN): (1) the \textbf{C}onvolutional \textbf{D}ensity \textbf{N}etwork of \textbf{G}aussian \textbf{M}ixtures (\textbf{CDN-GM}) for the task of generalistically modeling backgrounds; and (2) the \textbf{M}otion \textbf{E}stimation with \textbf{D}ifferencing \textbf{A}pproximation via \textbf{L}earning on a convolutional network (\textbf{MEDAL-net}) for context-driven foreground extraction. Our contributions are summarized:

\begin{itemize}
\item First, by leveraging existing technologies and being inspired by Bishop \cite{Bishop1995}, we present our CDN-GM, a feed-forward, highly parallelizable CNN that simulates a conditional probability density function of the temporal history at each pixel. Under this architecture, the network approximates a GMM statistical mapping function to produce models of their underlying multi-modular data distributions at each pixel. 
From the statistical models of pixel-wise data, backgrounds are extracted from the most informative components, resulting in an architecture that is light-weighted, compressed, and efficient.

\item Second, to model the underlying generator of input data, we propose a loss function based on unsupervised learning. This loss function directs CDN-GM to approximate the parameters of GMM conditioned on the data with expectation maximization. The resulting inferences will be made up of mixtures of Gaussian components that describe the data, and the most likely background description of actual observed data can be made on various inputs after training. In conjunction with CDN-GM, the proposed background modelling architecture achieves higher degrees of interpretability compared to the prior concept of estimating an implicit hidden function in neural network methods, and we gain better capability of adaptation to contextual dynamics as unsupervised learning allow training from an inexhaustible amount of data for learning with expectation maximization target.

\item Third, we design a compact convolutional auto-encoder for context-driven foreground extraction called MEDAL-net, which simulates a context-driven difference mapping between input frames and their corresponding background estimations. We take advantage of information found via background modeling, which includes both true background and corrupted (less observable, often by stopped objects) estimations, to capture semantic dynamics of scenes for motion detection. The network is trained via supervised learning in such a way that maintains good generalization across a scenario's dynamics.

\end{itemize}

The organization of this paper is as follows: Section \ref{section:related-works} encapsulates the synthesis of recent approaches in background initialization and foreground segmentation. The proposed method is described in Section \ref{section:proposed-method}. Experimental evaluations are discussed in Section \ref{section:experiments-and-discussion}. Finally, our conclusion and motivations towards future works are reached in Section \ref{section:conclusion}.

\begin{figure*}[!b]
	\vspace{-3mm}
	\centering
	\includegraphics[width=0.87\textwidth]{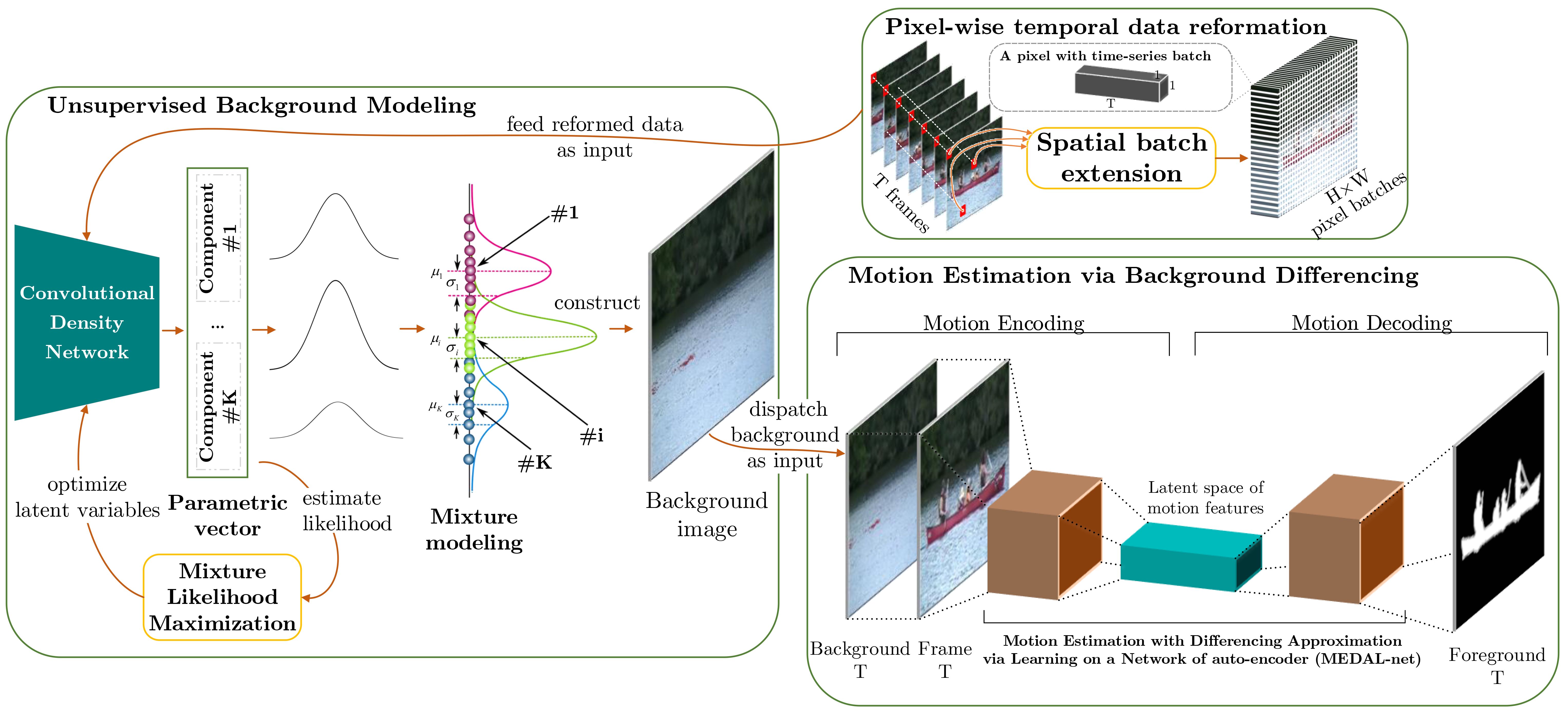}
	\caption{The overview of the proposed method for background modeling and foreground detection}
	\label{figure:01}
	\vspace{-5mm}
\end{figure*}

\section{Related Works}
\label{section:related-works}
Prior studies in recent decades were encapsulated in various perspectives of feature concepts \cite{Bouwmans2014a, Bouwmans2019, Garcia-Garcia2020}. Among published methods that meet the requirements of robustness, adaptation to scene dynamics, memory efficiency, and real-time processing, two promising approaches of background subtraction are statistical methods and neural-network-based models. Statistical studies aim to characterize the history of pixels' intensities with a probabilistic model. On the other hand, neural-network-based approaches implicitly estimate a non-linear mapping between an input sequence of observed scenes and hand-labeled background/foreground images.

In statistical approaches, the pixels' visual features are modeled with an explainable probabilistic foundation regarding either pixel-level or region-level in temporal and spatial resolution perspectives. In the last decades, there have been a variety of statistical models that were proposed to resolve the problem of background initialization. Stauffer and Grimson \cite{Stauffer1999} proposed a pioneering work that handled gradual changes in outdoor scenes using pixel-level GMM with a sequential K-means distribution matching algorithm. To enhance the foreground/background discrimination ability regarding scene dynamics, Pulgarin-Giraldo \textit{et al.} \cite{PulgarinGiraldo2017} improved GMM with a contextual sensitivity that used a Least Mean Squares formulation to update the parameter estimation framework. Validating the robustness of background modeling in a high amount of dynamic scene changes, Ha \textit{et al.} \cite{Synh2018} proposed a GMM with high variation removal module using entropy estimation. To enhance the performance, Lu \textit{et al.} \cite{Lu2018} applied a median filter on an input frame to reduce its spatial dimension before initializing its background. 
Overall, statistical models were developed with explicit probabilistic hypotheses to present the correlation of history observation at each image point or a pixel block, added with a global thresholding approach to extract foreground. This global thresholding technique for foreground detection usually leads to a compromise between the segregation of slow-moving objects and rapid adaptation to sudden scene changes within short-term measurement. This trade-off usually damages the image-background subtraction in multi-contextual scenarios, which is considered as a sensitive concern in motion estimation. Hence, regarding foreground segmentation from background modeling, it is critical to improve frame differencing from constructed background scenes with a better approximation mechanism.

Recently, there have also been many attempts to apply DNNs into background subtraction and background modeling problems with supervised learning. 
One of the earliest efforts to subtract the background from the input image frame was done by Braham \textit{et al.} \cite{Braham2016}. This work explores the potential of visual features learned by hidden layers for foreground-background pixel classification. Similarly, Wang \textit{et al.} \cite{Wang2017} proposed a deep CNN trained on only a small subset of frames as there is a large redundancy in a video taken by surveillance systems. The model requires a hand-labeled segmentation of moving regions as an indicator in observed scenes. Inspired from VGG-16 \cite{Simonyan2015}, Lim \textit{et al.} \cite{Lim2017} constructed an encoder-decoder architecture. The proposed encoder-decoder network takes a video frame, along its corresponding grayscale background and its previous frame as the network's inputs to compute their latent representations, and to deconvolve these latent features into a foreground binary map. Another method is DeepBS \cite{Babaee2018} which was proposed by Babaee \textit{et al.} to compute the background model using both SuBSENSE \cite{Charles2015} and Flux Tensor method \cite{Wang2014a}. The authors extract the foreground mask from a small patch from the current video frame and its corresponding background to feed into the CNN, and the mask is later post-processed to give the result. Nguyen \textit{et al.} proposed a motion feature network \cite{Nguyen2019} to exploit motion patterns via encoding motion features from small samples of images. The method's experimental results showed that the network obtained a promising results and well-performed on unseen data sequences. Recently, there is also a work from Chen \textit{et al.} \cite{Chen2019} which aims to exploit high-level spatial-temporal features with a deep pixel-wise attention mechanism and convolutional long short-term memory (ConvLSTM).

All things considered, most neural-network-based methods are benefitted from a significant number of weak statistical regularities in associative mapping, where the aim is to learn a transformation from an input batch of consecutive frames to the target hand-labeled foreground or background. There is little evidence that this supervised DNNs is ensured to possess true properties of observed scenes from the sampling peculiarities of training datasets, and be able to generalize to varying degree of contextual dynamics in the real-world. Furthermore, recent CNN methods do not ensure real-time performance, which is a crucial requirement for any practical systems. However, CNN's ability to utilize the parallelism mechanism of modern hardware efficiently, and the effective use of data for high-accuracy prediction is appealing to investigate. 

\section{The Proposed Method}
\label{section:proposed-method}
As shown with an overview in Fig. \ref{figure:01}, the primary goal of our proposed framework is to address the previously listed problems of DNNs and statistical methods, via adaptively acquiring the underlying properties of a sequence of images to construct corresponding background scenes with CDN-GM (the left subfigure), and extract foregrounds of interest through data-driven learning with MEDAL-net (the right lower subfigure). Following pixel-wise temporal data reformation (the right upper subfigure), a batch of video frames is decompressed into a sequence of pixel histories to estimate each pixel's true background intensity with CDN-GM. After reconstructing the background image from the output intensity sequence of CDN-GM, the input frame is concatenated along the channel dimension with the background to estimate the final segmentation map. The foreground extraction provided within MEDAL-net limits parameter search space and engenders context-driven difference mapping, rather than memorizing the single-valued mapping between input frames and labeled foregrounds.

\subsection{Convolution Density Network of Gaussian Mixture}
\label{subsection:convolution-density-network-of-gaussian-mixture}

According to Zivkovic's study \cite{Zivkovic2004}, let $\boldsymbol{\chi}_c^T = \{ {{\bf{x}}_1},{{\bf{x}}_2},...,{{\bf{x}}_T}|{{\bf{x}}_i} \in {[0,255]^c}\}$ be the time series of the $T$ most recently observed color signals of a pixel where the dimension of the vector ${{\bf{x}}_i}$ in the color space is $c$, the distribution of pixel intensity ${{\bf{x}}_i}$ can be modeled by a linear combination of $K$ probabilistic components $\boldsymbol\theta_k$ and their corresponding conditional probability density functions $P({{{\bf{x}}_i}}|{\boldsymbol\theta_k})$. The marginal probability $P({{{\bf{x}}_i}})$ of the mixture is defined in:

\begin{equation}
\label{eq:marginal-probability}
P({{\bf{x}}}) = \sum\limits_{k = 1}^K {P({\boldsymbol\theta_k})} P({{\bf{x}}}|{\boldsymbol\theta_k}) = \sum\limits_{k = 1}^K {{\pi _k}} P({{\bf{x}}}|{\boldsymbol\theta_k})
\end{equation}

\noindent where ${{\pi _k}}$ is the non-negative mixing coefficient that sums to unity, representing the likelihood of occurrence of the probabilistic component ${\boldsymbol\theta_k}$.

Because of the multimodality of observed scenes, the intensity of target pixels is assumed to be distributed normally in a finite mixture. Regarding RGB space of analyzed videos, each examined color channel in $\textbf{x}_i$ was assumed to be distributed independently and can be described with a common variance $\sigma _k$ to avoid performing costly matrix inversion as indicated in \cite{Stauffer1999}. Hence, the multivariate Gaussian distribution can be re-formulated as:

\begin{equation}
\label{eq:simple-multivariate-gaussian-mixture}
\begin{aligned}
P({{\bf{x}}}|{\boldsymbol\theta_k}) &= \mathcal{N}({{\bf{x}}}|{\boldsymbol\mu _k},{\sigma_k}) \\ 
&= \frac{1}{{\sqrt {{{(2\pi )}^c}{\sigma_k^c}} }}\exp \left( { - \frac{\parallel {{\bf{x}}} - {{\boldsymbol{\mu }}_k}{\parallel ^2}}{2\sigma_k}} \right)
\end{aligned}
\end{equation}

\noindent where ${\boldsymbol\mu _k}$ is the estimated mean and $\sigma _k$ is the estimated universal covariance of examined color channels in the $k^{th}$ Gaussian component.

From this hypothesis, in this work, we propose an architecture of convolutional neural network, called Convolutional Density Network of Gaussian Mixtures (CDN-GM), which employs a set of non-linearity transformations ${f_\theta }\left(  \cdot  \right)$ to formulate a conditional formalism of GMM density function of $\bf{x}$ given a set of randomly selected, vectorized data points $\boldsymbol{\chi}_T$:

\begin{equation}
\label{eq:formulate-equation}
{\bf{y}}_T = {f_\theta } (\boldsymbol{\chi}^T_c) \sim P({{\bf{x}|\boldsymbol{\chi}^T_c}})
\end{equation}

The ability of multilayer neural networks that was trained with an optimization algorithm to learn complex, high-dimensional, nonlinear mappings from large collections of examples increases their capability in pattern recognition via gathering relevant information from the input and eliminating irrelevant variabilities. With respect to problems of prediction, the conditional average represents only a very limited statistic. For applicable contexts, it is considerably beneficial to obtain a complete description of the probability distribution of the target data. In this work, we incorporate the mixture density model with the convolutional neural network instead of a multi-layer perceptron as done by Bishop \textit{et al.} in the vanilla research \cite{Bishop1995}. In the proposed scheme, the network itself learns to act as a feature extractor to formulate statistical inferences on temporal series of intensity values. Literally, a background image contains most frequently presented intensities in the sequence of observed scenes. Hence, in CDN-GM, we take advantage of this mechanism to exploit the most likely intensity value that will raise in the background image via consideration of temporal arrangement. The scheme of weight sharing in the proposed CNN reduces the number of parameters, making CDN-GM lighter and exploiting the parallel processing of a set of multiple pixel-wise analysis within a batch of frames.

\begin{figure*}[!t]
	\vspace{-5mm}
	\centering
	\includegraphics[width=0.8\textwidth]{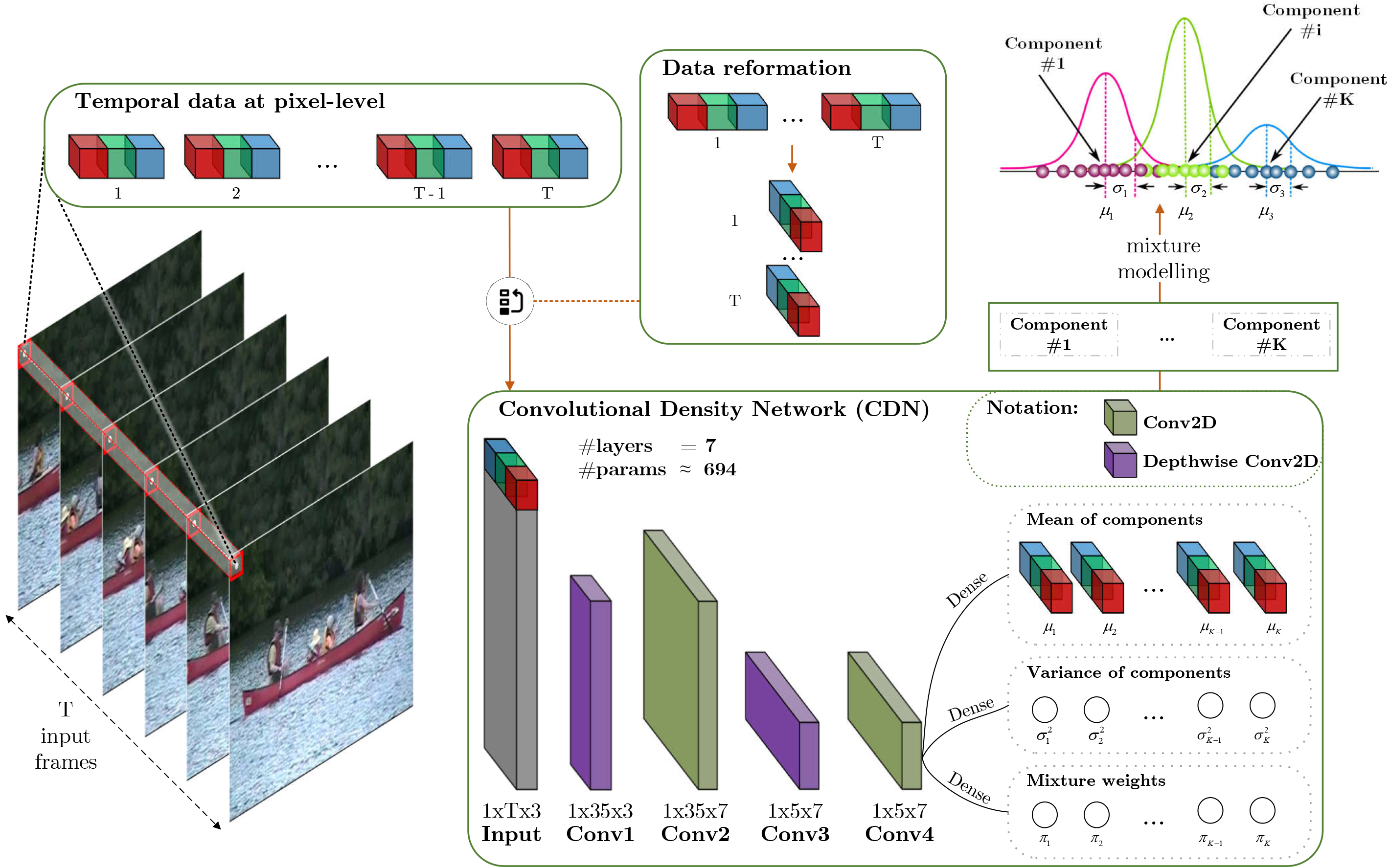}
	\caption{The proposed architecture of Convolution Density Network of Gaussian Mixture Model}
	\label{figure:02}
	\vspace{-3mm}
\end{figure*}

The main goal of CDN-GM is to construct an architecture of CNN which presents multivariate mapping in forms of Gaussian Mixture Model with the mechanism of offline learning. With the simulated probabilistic function, we aim to model the description of the most likely background scenes from actual observed data. In other words, the regularities in the proposed CNN should cover a generalized presentation of the intensity series of a set of consecutive frames at pixel level. To achieve this proposition, instead of using separate GMM for each pixel-wise statistical learning, we consider to use a single GMM to formulate the temporal history of all pixels in the whole image. Accordingly, CDN-GM architecture is extended through a spatial extension of temporal data at image points. The network architecture contains seven learned layers, not counting the input -- two depthwise convolutional, two convolutional and three dense layers as summarized in Fig. \ref{figure:02}.

The network output ${\bf{y}_T}$, whose dimension is $\left( {c + 2} \right) \times K$, is partitioned into three portions ${\bf{y}}_\mu \left( \boldsymbol{\chi}^T_c \right)$, ${\bf{y}}_\sigma \left( \boldsymbol{\chi}^T_c \right)$, and ${\bf{y}}_\pi \left( \boldsymbol{\chi}^T_c \right)$ corresponding to the latent variables of GMM:

\begin{equation}
\label{eq:output-formulate}
\begin{aligned}
{\bf{y}_T} &= [{\bf{y}}_\mu \left( \boldsymbol{\chi}^T_c \right), {\bf{y}}_\sigma \left( \boldsymbol{\chi}^T_c \right), {\bf{y}}_\pi \left( \boldsymbol{\chi}^T_c \right)] \\
&= [{\bf{y}}_\mu ^1, \ldots ,{\bf{y}}_\mu ^K,{\bf{y}}_\sigma ^1, \ldots ,{\bf{y}}_\sigma ^K,{\bf{y}}_\pi ^1, \ldots ,{\bf{y}}_\pi ^K]
\end{aligned}
\end{equation}

With our goal of formulating the GMM, we impose a different restriction on threefold outputs from the network:

\begin{itemize}
\item First, as the mixing coefficients $\pi_k$ indicate the proportion of data accounted for by mixture component $k$, they must be defined as independent and identically distributed probabilities. To achieve this regulation, in principle, we activate the network output with a softmax function:

\begin{equation} 
\label{eq:softmax-function}
\pi_k(\boldsymbol{\chi}^T_c) = \frac{{\exp ({\bf{y}}_\pi ^k)}}{{\sum\nolimits_{l = 1}^K {\exp ({\bf{y}}_\pi ^l)} }}
\end{equation}

\item Second, in the realistic scenarios, the measured intensity of observed image signals may fluctuate due to a variety of factors, including illumination transformations, dynamic contexts and bootstrapping. In order to conserve the estimated background, we have to restrict the value of the variance of each component to the range $[\bar \sigma_{min}, \bar \sigma_{max}]$ so that each component does not span spread the entire color space, and does not focus on one single color cluster:

\begin{equation}
\label{eq:variance-constraint}
\sigma _k (\boldsymbol{\chi}^T_c) = \frac{\bar\sigma_{min} \times (1-\hat\sigma_k) + \bar\sigma_{max} \times \hat\sigma_k }{255}
\end{equation}

\noindent where $\sigma _k (\boldsymbol{\chi}^T_c)$ is normalized towards a range of $[0,1]$ over the maximum color intensity value, 255; and $\hat\sigma_{k}$ is the normalized variance that was activated through a hard-sigmoid function from the output neurons ${\bf{y}}_\sigma $ that correspond to the variances:

\begin{equation}
\label{eq:hard-sigmoid-sigma}
{{\hat \sigma }_k}(\boldsymbol{\chi}^T_c) = \max \left[ {0,\min \left( {1,\frac{{{\bf{y}}_\sigma ^k + 1}}{2}} \right)} \right]
\end{equation}

\noindent In this work, we adopt the hard sigmoid function because of the piecewise linear property and correspondence to the bounded form of a linear rectifier function (ReLU) of the technique. Furthermore, this was proposed and proved to be more efficient in both in software and specialized hardware implementations by Courbariaux \textit{et al.} \cite{Courbariaux2015}.

\item Third, the mean of the probabilistic mixture is considered on a normalized RGB color space where the intensity values retain in a range of $[0,1]$ so that they can be approximated correspondingly with the normalized input. Similar to the normalized variance $\hat\sigma_{k}$, the mixture mean is standardized from the corresponding network outputs with a hard-sigmoid function:

\begin{equation}
\label{eq:hard-sigmoid-mu}
\mu_k (\boldsymbol{\chi}^T_c) = \max \left[ {0,\min \left( {1,\frac{{{\bf{y}}_\mu ^k + 1}}{2}} \right)} \right]
\end{equation}

\end{itemize}

From the proposed CNN, we extract the periodical background image for each block of pixel-wise time series of data in a period of $T$. This can be done by selecting the means whose corresponding distributions have the highest degree of high-weighted, low-spread. To have a good grasp of the importance of a component in the mixture, we use a different treatment of weight updates with a ratio of ${{{\pi _{k'}}({\boldsymbol{\chi }}_c^T)} \mathord{\left/ {\vphantom {{{\pi _{k'}}({\boldsymbol{\chi }}_c^T)} {{\sigma _{k'}}({\boldsymbol{\chi }}_c^T)}}} \right. \kern-\nulldelimiterspace} {{\sigma _{k'}}({\boldsymbol{\chi }}_c^T)}}$. This is the manner of weighting components within a mixture at each pixel by valuing high-weighted, low-spread distributions in the mixture, thereby spotlighting the most significant distribution contributing to the construction of backgrounds.

\begin{equation} 
\label{equation:13}
BG(\boldsymbol{\chi}^T_c) = \mathop {\max } (\mu_{k}  \cdot \hat {BG}_{k,T})\text{,}
\quad \text{for } k \in \left[ {1,K} \right]
\end{equation}

where background mapping is defined at each pixel $\textbf{x}$ as:

\begin{equation} 
\label{equation:14}
{{\hat {BG}}_{k,T}}({\boldsymbol{\chi}}^T_c) = \left\{ 
\begin{array}{l}
	\begin{multlined}
	1, \quad \text{if } \mathop {argmax}\limits_{k'} [{\raise0.7ex\hbox{${{\pi _{k'} ({\boldsymbol{\chi}}^T_c)}}$} \!\mathord{\left/
 {\vphantom {{{\pi _k }} {{\sigma _k}}}}\right.\kern-\nulldelimiterspace}
\!\lower0.7ex\hbox{${{\sigma _{k'} ({\boldsymbol{\chi}}^T_c)}}$}}] = k \\
	\text{for } k \in \left[ {1,K} \right]
	\end{multlined}	\\
	0, \quad \text{otherwise}
\end{array}\right.\\ 
\end{equation}

\subsection{The unsupervised loss function of CDN-GM}
\label{subsection:loss-function}
In practice, particularly in each real-life scenario, the background model must capture multiple degrees of dynamics, which is more challenging by the fact that scene dynamics may also change gradually under external effects (e.g. lighting deviations). These effects convey the latest information regarding contextual deviations that may constitute new background predictions. Therefore, the modeling of backgrounds must not only take into account the various degrees of dynamics across multiple imaging pixels of the data source, but it must also be able to adaptively update its predictions with respect to semantic changes. Equivalently, in order to approximate a statistical mapping function for background modeling, the proposed neural network function has to be capable of approximating a conditional probability density function, thereby estimating a multi-modular distribution conditioned on its time-wise latest raw imaging inputs. The criteria for the neural statistical function to be instituted can be summarized as follows:

\begin{itemize}
\item As a metric for estimating distributions, input data sequences cannot be weighted in terms of order.

\item Taking adaptiveness into account, the neural probabilistic density function can continuously interpolate predictions in evolving scenes upon reception of new data.

\item The neural network function has to be generalizable such that its model parameters are not dependent on specific learning datasets.
\end{itemize}

Hence, satisfying the prescribed criteria, we propose a powerful loss function capable of directing the model's parameters towards adaptively capturing the conditional distribution of data inputs, thereby approximating a statistical mapping function in a technologically parallelizable form. At every single pixel, the proposed CNN estimates the probabilistic density function on the provided data using its GMM parameters. Specifically, given the set $\boldsymbol{\chi}^T_c$ randomly selected, vectorized data points, it is possible to retrieve the continuous conditional distribution of the data target $\bf{x}$ with the following functions:

\begin{equation}
\label{eq:sss}
P({{\bf{x}}}) = \sum\limits_{k = 1}^K {{\pi _k}}({\boldsymbol{\chi}^T_c}) \cdot \mathcal{N}({{\bf{x}}}|{\boldsymbol\mu _k},{\sigma_k})
\end{equation}

\noindent where the general disposition of this distribution is approximated by a finite mixture of Gaussians, whose values are dependent on our learnable neural variables:

\begin{equation}
\label{eq:}
\mathcal{N}({{\bf{x}}}|{\boldsymbol\mu _k},{\sigma_k}) = {1 \over {\sqrt {2\pi  \cdot {\sigma_k}{{({\boldsymbol{\chi}^T_c})}^2}} }} \cdot \exp \left\{ { - {{{{\left\| {{{\bf{x}}} - {\mu _k}({\boldsymbol{\chi}^T_c})} \right\|}^2}} \over {2{\sigma_k}{{({\boldsymbol{\chi}^T_c})}^2}}}} \right\}
\end{equation}

In our proposed loss function, the data distribution to be approximated is the set of data points relevant to background construction. This is rationalized by the proposed loss function's purpose, which is to direct the neural network's variables towards generalizing universal statistical mapping functions. Furthermore, even with constantly evolving scenes where the batches of data values also vary, this loss measure can constitute fair weighting on the sequence of inputs. Our proposed loss measure is designed to capture various pixel-wise dynamics over a video scene and to encompass even unseen perspectives via exploiting the huge coverage of multiple scenarios across more than one case with data. In other words, the order of the network's input does not matter upon loading, which is proper for any statistical function on estimating distribution. For modeling tasks, we seek to establish a universal multi-modular statistical mapping function on the RGB color space, which would require optimizing the loss not just on any single pixel, but for $b$ block of time-series image intensity data fairly into a summation value.

\begin{equation}
\vspace{-2mm}
\label{eq:los-funct}
{\cal L} = \sum\limits_i^b {\sum\limits_j^T {\cal L}_{ j}^{(i)} } 
\end{equation}

\begin{equation}
\label{eq:log-of-mixtures}
\textnormal{where}\quad\quad\quad{\cal L}_{ j}^{(i)} = - \ln \left( {\sum\limits_{k = 1}^K {\pi _k^{(i)}} {\cal N}({{\bf{x}}_j}|{\bf{\mu }}_k^{(i)},\sigma _k^{(i)})} \right)
\end{equation}

\noindent where ${\bf{x}}_j$ is the $j^{th}$ element of the $i^{th}$ time-series data ${\boldsymbol{\chi}^{T,(i)}_c}$ of pixel values; $\pi^{(i)}$, $\mu ^{(i)}$, and $\sigma ^{(i)}$ are respectively the desired mixing coefficients, means, and variances that commonly model the distribution of ${\boldsymbol{\chi}^{T,(i)}_c}$ in GMM.

We define ${\cal L}_{ j}^{(i)}$ as the error function for our learned estimation on an observed data point ${\bf{x}}_j$, given the locally relevant dataset ${\boldsymbol{\chi}^{T,(i)}_c}$ for the neural function, \textit{i.e.} CDN-GM. ${\cal L}_{ j}^{(i)}$ is based on the statistical log-likelihood function and is equal to the negative of its magnitude. Hence, by minimizing this loss measure, we will essentially be maximizing the expectation value of the GMM-based neural probabilistic density function $P({{\bf{x}}})$, from the history of pixel intensities at a pixel position. Employing stochastic gradient descent on the negative logarithmic function ${\cal L}_{ j}^{(i)}$ involves not only monotonic decreases, which are steep when close to zero, but also upon convergence it also leads to the proposed neural function approaching an optimized mixture of Gaussians probability density function.

In addition, because the objective of our network is to maximize the likelihood of the output on the data itself, without any external labels, to estimate background scenes, the proposed work can be considered an unsupervised approach.
The key thing here is that whether the neural network can learn to optimize the loss function with the standard stochastic gradient descent algorithm with \textit{back-propagation}. This can only be achieved if we can obtain suitable equations of the partial derivatives of the error $\mathcal{L}$ with respect to the outputs of the network. As we describe in the previous section, ${\bf{y}}_\mu $, ${\bf{y}}_\sigma $, and ${\bf{y}}_\pi $ present the proposed CDN-GM's outputs that formulate to the latent variables of GMM. The partial derivative ${{\partial \mathcal{L}_j^{(i)}} \mathord{\left/ {\vphantom {{\partial \mathcal{L}_j^{(i)}} {\partial {o_k}}}} \right. \kern-\nulldelimiterspace} {\partial {{\bf{y}}^{(k)}}}}$ can be evaluated for a particular pattern and then summed up to produce the derivative of the error function $\cal L$. To simplify the further analysis of the derivatives, it is convenient to introduce the following notation that presents the posterior probabilities of the component $k$ in the mixture, using Bayes theorem:

\begin{equation}
\label{eq:convenient-notation}
\Pi _k^{(i)} = \frac{{\pi _k^{(i)}\mathcal{N}({{\mathbf{x}}_j}|{{\boldsymbol\mu }}_k^{(i)},\sigma _k^{(i)})}}{{\sum\limits_{l = 1}^K {\pi _l^{(i)}} \mathcal{N}({{\mathbf{x}}_j}|{{\boldsymbol\mu }}_l^{(i)},\sigma _l^{(i)})}}
\end{equation}

First, we need to consider the derivatives of the loss function with respect to the network's outputs ${\bf{y}}_{\pi}$ that correspond to the mixing coefficients $\pi_k$. Using Eq. (\ref{eq:log-of-mixtures}) and (\ref{eq:convenient-notation}), we obtain:

\begin{equation}
\label{eq:partial-deriv-loss-wrt-mixture}
\frac{{\partial \mathcal{L}_j^{(i)}}}{{\partial \pi _k^{(i)}}} = \frac{{\Pi _k^{(i)}}}{{\pi _k^{(i)}}}
\end{equation}

\noindent From this expression, we perceive that the value of $\pi_k^{(i)}$ explicitly depends on ${\bf{y}}^{(l)}_{\pi}$ for $l=1,2,...,K$ as $\pi_k^{(i)}$ is the result of the softmax mapping from ${\bf{y}}^{(l)}_{\pi}$ as indicated in Eq. (\ref{eq:softmax-function}). We continue to examine the partial derivative of $\pi_k^{(i)}$ with respect to a particular network output ${\bf{y}}^{(l)}_{\pi}$, which is

\begin{equation}
\label{eq:mixture-softmax-deriv}
\frac{{\partial \pi _k^{(i)}}}{{\partial {\bf{y}}^{(l)}_{\pi} }} = \left\{ {\begin{array}{*{20}{l}}
  {\pi _k^{(i)}(1 - \pi _l^{(i)}{\text{),}}}&{\quad {\text{if }}k = l} \\ 
  { - \pi _l^{(i)}\pi _k^{(i)}{\text{,}}}&{\quad {\text{otherwise}}{\text{.}}} 
\end{array}} \right.
\end{equation}

\noindent By chain rule, we have

\begin{equation}
\label{eq:loss-wrt-output-chain-rule-deriv}
\frac{{\partial \mathcal{L}_j^{(i)}}}{{\partial {\bf{y}}^{(l)}_{\pi} }} = \sum\limits_k {\frac{{\partial \mathcal{L}_j^{(i)}}}{{\partial \pi _k^{(i)}}}} \frac{{\partial \pi _k^{(i)}}}{{\partial {\bf{y}}^{(l)}_{\pi} }}
\end{equation}

\noindent From Eq. (\ref{eq:convenient-notation}), (\ref{eq:partial-deriv-loss-wrt-mixture}), (\ref{eq:mixture-softmax-deriv}), and (\ref{eq:loss-wrt-output-chain-rule-deriv}), we then obtain

\begin{equation}
\label{eq:loss-wrt-mixture-output-unit}
\frac{{\partial \mathcal{L}_j^{(i)}}}{{\partial \bf{y}}^{(l)}_{\pi}} = \pi _l^{(i)} - \Pi _l^{(i)}
\end{equation}

For ${\bf{y}}^{(k)}_{\sigma}$, we make use of Eq. (\ref{eq:simple-multivariate-gaussian-mixture}), (\ref{eq:variance-constraint}), (\ref{eq:hard-sigmoid-sigma}), (\ref{eq:log-of-mixtures}), and (\ref{eq:convenient-notation}), by differentiation, to obtain

\begin{equation}
\label{eq:derivative-of-var-out}
\frac{{\partial \mathcal{L}_j^{(i)}}}{{\partial {\bf{y}}^{(k)}_{\sigma} }} = \frac{{3.2}}{{255}}\Pi _k^{(i)}\left( {\frac{c}{2}\sqrt {{{(2\pi )}^c}{{(\sigma _k^{(i)})}^{c + 2}}}  - \frac{{\parallel {{\mathbf{x}}_j} - {{\boldsymbol{\mu }}_k}{\parallel ^2}}}{{2{{(2\pi )}^c}{{(\sigma _k^{(i)})}^{c + 2}}}}} \right)
\end{equation}

for $-2.5 < {\bf{y}}^{(k)}_{\sigma} < 2.5$. This is because the piece-wise property in the definition of the hard-sigmoid function.

Finally, for ${\bf{y}}^{(k)}_{\mu}$, let $\mu_{k,l}^{(i)}$ be the $l^{th}$ element of the mean vector where $l$ is an integer lies in $[0,c)$ and suppose that $\mu_{k,l}^{(i)}$ corresponds to an output $o_k^\mu$ of the network. We can get derivative of $\mu_{k,l}^{(i)}$ by taking Eq. (\ref{eq:simple-multivariate-gaussian-mixture}), (\ref{eq:hard-sigmoid-mu}), (\ref{eq:log-of-mixtures}), (\ref{eq:convenient-notation}) into the differentiation process:

\begin{equation}
\label{eq:derivative-of-mix-out}
\frac{{\partial \mathcal{L}_j^{(i)}}}{{\partial {\bf{y}}^{(k)}_{\mu} }} = 0.2 \times \Pi _k^{(i)}\frac{{{x_{j,l}} - \mu _{k,l}^{(i)}}}{{\sigma _k^{(i)}}}
\end{equation}

\noindent for $-2.5 < {\bf{y}}^{(k)}_{\mu} < 2.5$.

From Eq. (\ref{eq:loss-wrt-mixture-output-unit}), (\ref{eq:derivative-of-var-out}), and (\ref{eq:derivative-of-mix-out}), when CDN-GM is performed data-driven learning individually on each video sequence using Adam optimizer with a learning rate of $\alpha$, the process tries to regulate the values of laten parameters in the mixture model via minimizing the negative of log likelihood function. Hence, once the proposed model has been trained on video sequences, it is obviously seen that the network can predict the conditional density function of the target background, which is a statistical description of time-series data of each image point, rather than memorizing the single-valued mapping between input frames and labelled backgrounds.


\begin{figure*}[!b]
	\vspace{-3mm}
	\centering
	\includegraphics[width=0.85\textwidth]{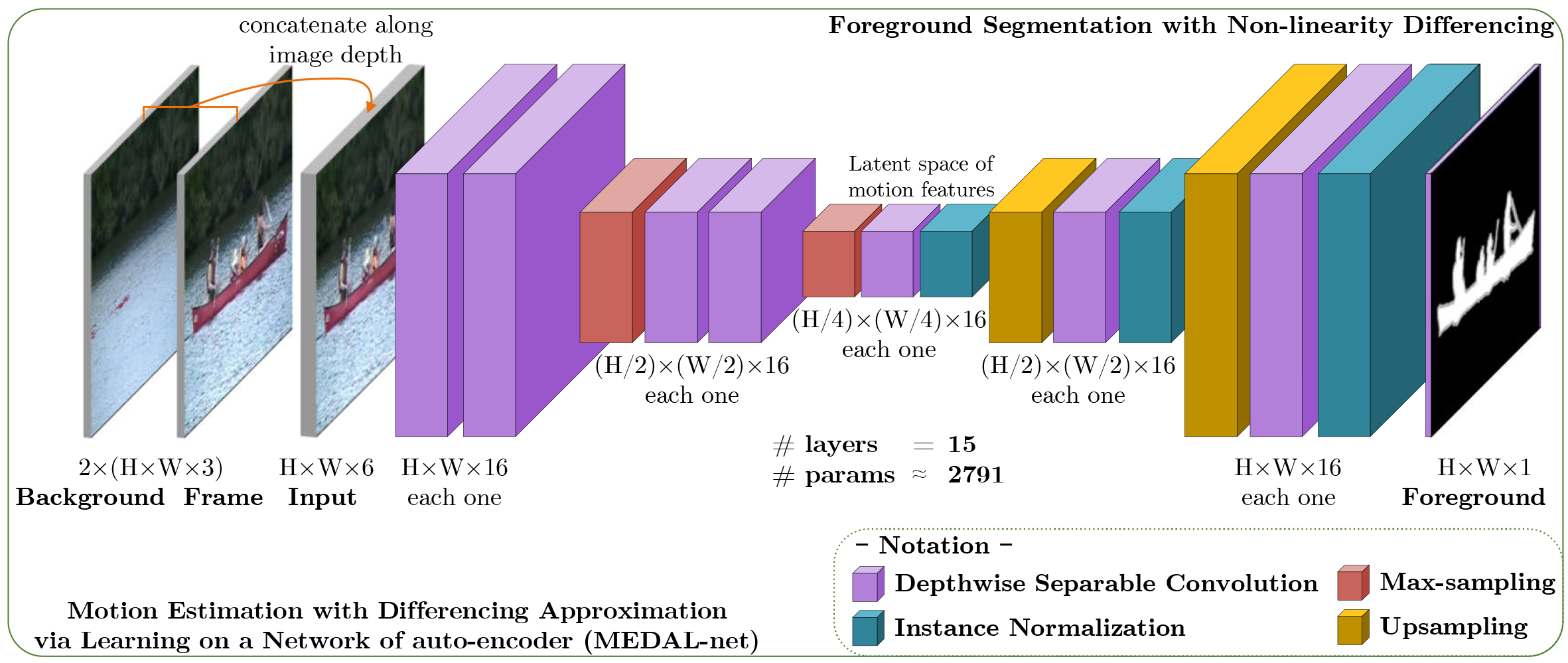}
	\caption{The proposed architecture of MEDAL-net grounded on convolutional autoencoder for foreground detection}
	\label{figure:03}
	\vspace{-5mm}
\end{figure*}

\subsection{Foreground Segmentation with Non-linear Differencing}
\label{subsection:foreground-segmentation}
In this section, we present the description of our proposed convolutional auto-encoder, called MEDAL-net, which simulates non-linear frame-background differencing for foreground detection. Traditionally, thresholding schemes are employed to find the highlighted difference between an imaging input and its corresponding static view in order to segment motion. For example, Stauffer and Grimson \cite{Stauffer1999} employed variance thresholding on background - input pairs by modeling the static view with GMM. While experimental results suggest certain degrees of applicability due to its simplicity, the approach lacks in flexibility as the background model is usually not static and may contain various motion effects such as occlusions, stopped objects, shadow effects, etc. 

In practice, a good design of a difference function between the current frame and its background must be capable of facilitating motion segmentation across a plethora of scenarios and effects. However, regarding countless scenarios in real life, where there are unique image features and motion behaviors to each, there is yet any explicit mathematical model that is general enough to cover them all. Because effective subtraction requires high-degreed non-linearity in order to compose a model for the underlying mathematical framework of many scenarios, following the Universal approximation theorem \cite{Hornik1989}, we design the technologically parallelizable neural function for an approximation of such framework. Specifically, we make use of a CNN to construct a foreground segmentation network. The motive is further complemented by two folds:

\begin{itemize}
\item Convolutional Neural Networks have long been known for their effectiveness in approximating nonlinear functions with arbitrary accuracy.

\item Convolutional Neural Networks are capable of balancing between both speed and generalization accuracy, especially when given an effective design and enough representative training data.
\end{itemize}

However, recent works exploiting CNN in motion estimation are still generating heavy-weighted models which are computationally expensive and not suitable for real-world deployment. In our proposed work, we exploit the use of a pair of the current video frame and its corresponding background as the input to the neural function and extract motion estimation. By combining this with a suitable learning objective, we explicitly provide the neural function with enough information to mold itself into a context-driven non-linear difference function, thereby restricting model behavior and its search directions. This also allows us to scale down the network’s parameter size, width, and depth to focus on learning representations while maintaining generalization for unseen cases. As empirically shown in the experiments, the proposed architecture is light-weighted in terms of the number of parameters, and is also extremely resource-efficient, e.g. compared to FgSegNet \cite{Lim2018}.

\subsubsection{Architectural design}
The overall flow of the MEDAL-net is shown in Fig. \ref{figure:03}. We employ the encoder-decoder design approach for our segmentation function. With this approach, data inputs are compressed into a low-dimensional latent space of learned informative variables in the encoder, and the encoded feature map is then passed into the decoder, thereby generating foreground masks. 

In our design, we fully utilize the use of depthwise separable convolution introduced in MobileNets \cite{Howard2017} so that our method can be suitable for mobile vision applications. Because this type of layer significantly scales down the number of convolutional parameters, we reduced the number of parameters of our network by approximately 81.7\% compared to using only standard 2D convolution, rendering a light-weighted network of around 2,800 parameters. Interestingly, even with such a small set of parameters, the network still does not lose its ability to generalize predictions at high accuracy. Our architecture also employs normalization layers, but only for the decoder. This design choice is to avoid the loss of information in projecting the contextual differences of background-input pairs into the latent space via the encoder, while formulating normalization to boost the decoder’s learning.

\paragraph{Encoder}
The encoder can be thought of as a folding function that projects the loaded data into an information-rich low-dimensional feature space. In our architecture, the encoder takes in pairs of video frames and their corresponding backgrounds concatenated along the depth dimension as its inputs. Specifically, the background image estimated by CDN-GM is concatenated with imaging signals such that raw information can be preserved for the neural network to freely learn to manipulate. Moreover, with the background image also in its raw form, context-specific scene dynamics (e.g. moving waves, camera jittering, intermittent objects) are also captured. Thus, as backgrounds are combined with input images to formulate predictions, MEDAL-net may further learn to recognize motions that are innate to a scene, thereby selectively segmenting motions of interest based on the context.

In addition, by explicitly providing a pair of the current input frame and its background image to segment foregrounds, our designed network essentially constructs a simple difference function that is capable of extending its behaviors to accommodate contextual effects. Thus, we theorize that approximating this neural difference function would not require an enormous number of parameters. In other words, it is possible to reduce the number of layers and the weights' size of the foreground extraction network to accomplish the task. Hence, the encoder only consists of a few convolutional layers, with 2 max-pooling layers for downsampling contextual attributes into a feature-rich latent space.

\paragraph{Decoder}
The decoder of our network serves to unfold the encoded feature map into the foreground space using convolutional layers with two upsampling layers to restore the original resolution of its input data. 

In order to facilitate faster training and better estimation of the final output, we engineered the decoder to include instance normalization, which is apparently more efficient than batch normalization \cite{Ulyanov2017}. Using upsampling to essentially expand the latent tensors, the decoder also employs convolutional layers to induce non-linearity like the encoder.

The final output of the decoder is a grayscale probability map where each pixel’s value represents the chance that it is a component of a foreground object. This map is the learned motion segmentation results with pixel-wise confidence scores determined on account of its neighborhood and scene-specific variations. In our design, we use the hard sigmoid activation function because of its property that allows faster gradient propagation, which results in less training time.

At inference time, the final segmentation result is a binary image obtained by placing a constant threshold on the generated probability map. Specifically, suppose $\mathbf{X}$ is a probability map of size $N \times H \times W \times 1$, and let the set $F$ be defined as:

\begin{equation}
\label{eq:F-function}
F = \left\{ {(x,y,z)|{{\bf{X}}_{x,y,z,0}} \ge \epsilon} \right\}
\end{equation}

\noindent where $x \in [0,N]$, $y \in [0,H]$, $z \in [0,W]$, and $\epsilon$ is an experimentally determined parameter. In other words, $F$ is a set of indices of $\mathbf{X}$ that satisfy the threshold $\epsilon$. The segmentation map $\hat{\mathbf{Y}}$ of size $N \times H \times W$ is obtained by:

\begin{equation}
\label{eq:segmap-Yij}
\hat{\mathbf{Y}}_{i,j,k} = \begin{cases} 
      1, & (i,j,k) \in F \\
      0, & otherwise
   \end{cases}
\end{equation}

\noindent where 1 represents  indices classified as foreground, and 0 represents background indices.

\subsubsection{Model learning}
We penalize the output of the network using the cross-entropy loss function commonly used for segmentation tasks $\left[ {x,y,z} \right]$, as the goal of the model is to learn a Dirac delta function for each pixel. The description of the loss function is as follows:

\begin{equation}
\label{eq:Loss-MEDAL-net}
\begin{aligned}
L= - \frac{1}{N} \sum_{i=1}^{N}  \sum_{j=1}^{H}  \sum_{k=1}^{W}  [ & \mathbf{Y}_{i,j,k} \log(\hat{\mathbf{Y}}_{i,j,k}) \\
& + (1 - \mathbf{Y}_{i,j,k}) \log(1 - \hat{\mathbf{Y}}_{i,j,k})]
\end{aligned}
\end{equation}

\noindent where $\mathbf{Y}$ is the corresponding set of foreground binary masks for $\hat{\mathbf{Y}}$, the batch of predicted foreground probability maps. 

The designed architecture is enabled to learn not only pixel-wise motion estimates of the training set, but it also is taught to recognize inherent dynamics in its data, and perform as a context-driven neural difference function to accurately interpolate region-wise foreground predictions of unseen perspectives.

\section{Experiments and Discussion}
\label{section:experiments-and-discussion}

\subsection{Experimental Setup}
In this section, we verify experimentally the capabilities of the proposed CDN-MEDAL-net via comparative evaluations in background modeling and subtraction. Our proposed scheme is designed to 
compete with state-of-the-art works in accuracy and processing speed within a light-weighted structure. Therefore, we compare the accuracy of the proposed framework not only with unsupervised approaches that are light-weighted and generalizable without pretraining: GMM -- Stauffer \& Grimson \cite{Stauffer1999}, GMM -- Zivkovic \cite{Zivkovic2004}, SuBSENSE \cite{Charles2015}, PAWCS \cite{Charles2015a}, TensorMoG \cite{Ha2020}, BMOG \cite{Martins2018}, FTSG \cite{Wang2014a}, SWCD \cite{Sahin2018}, but also with the data-driven, supervised models which trade computational expenses for high accuracy performance: FgSegNet\_S \cite{Lim2018}, FgSegNet \cite{Lim_2018}, FgSegNet\_v2 \cite{Lim2019}, Cascade CNN \cite{Wang2017}, DeepBS \cite{Babaee2018}, STAM \cite{Liang2019}.

First, we proceed with an ablation study to evaluate the contribution of the background modelling of CDN-GM on the Scene Background Modeling (SBMnet) dataset \cite{Jodoin2017}. The metrics include AGE (Average Gray-level Error), pEPs (Percentage of Error Pixels), pCEPs (Percentage of Clustered Error Pixels), MS-SSIM (MultiScale Structural Similarity Index), PSNR (Peak-Signal-to-Noise-Ratio), and CQM (Color image Quality Measure). The first three measure the intensity-level error difference between the algorithm's output with the provided ground-truth as how methods are sensitive to small variations and require intensity-level exactness with referenced ground-truth, where lower estimation values indicate better background estimates. The latter three focus on quantifying visual and structural quality of generated backgrounds by algorithms, where higher values indicate better results.

In terms of background subtraction results, second, we employ quantitative analysis on the CDnet-2014 \cite{Wang2014} dataset. Our metrics are those that can be appraised from confusion matrices, i.e. Precision, Recall, F-Measure, False-Negative Rate (FNR), False-Positive Rate (FPR) and Percentage of Wrong Classification (PWC). The benchmarks were performed by comparing foreground predictions against provided ground-truths. Through our results, we observe the capabilities of MEDAL-net in leveraging background models of CDN-GM for context-driven background subtraction. 



In our experiments, the number of Gaussians $K$ is empirically and heuristically to balance the CDN-GM's capability of modeling constantly evolving contexts (e.g. moving body of water) under many effects of potentially corruptive noises.
With $K$ too big, many GMM components many be unused or they simply capture various noises within contextual dynamics. As the Gaussian component corresponding to the background intensity revolves around the most frequently occurring color subspaces to draw predictions, the other components serve only as either placeholders for abrupt changes in backgrounds, or capture intermittent noises of various degrees. 
Our proposed CDN-GM model was set up with the number of Gaussian components $K = 3$ for all experimented sequences.
In addition, the constants $\bar\sigma_{min}$ and $\bar\sigma_{max}$ were chosen such that no Gaussian components span the whole color space while not contracting to a single point that represents noises. 
If the $\left[ {{{\bar \sigma }_{min}},{{\bar \sigma }_{\max }}} \right]$ interval is too large, some of the components might still cover all intensity values, making it hard to find the true background intensity. From experimental observations, we find that the difference between color clusters usually does not exceed approximately $16$ at minimum and $32$ at maximum.


\begin{table}[!b]
\vspace{-3mm}
\setcounter{table}{4}
\caption{Comparison on the Scene Background Modeling (SBMnet) dataset.}
\label{tab:sbmnet-comparison}
\centering
\resizebox{0.49\textwidth}{!}{
\begin{tabular}{l|llllll}

\toprule
\multicolumn{1}{c|}{\textbf{Method}}            & \multicolumn{1}{c}{\textbf{AGE}} & \multicolumn{1}{c}{\textbf{pEPs}} & \multicolumn{1}{c}{\textbf{pCEPs}} & \multicolumn{1}{c}{\textbf{MS-SSIM}} & \multicolumn{1}{c}{\textbf{PSNR}} & \multicolumn{1}{c}{\textbf{CQM}} \\

\midrule
GMM -- S \& G & \textcolor{blue}{$15.7189_{(3)}$}  & \textcolor{red}{$0.1275_{(1)}$}   & \textcolor{blue}{$0.1018_{(3)}$}  & \textcolor{green}{$0.8639_{(2)}$} & \textcolor{red}{$26.1304_{(1)}$}   & \textcolor{red}{$26.9061_{(1)}$} \\
GMM -- Zivkovic            & \textcolor{red}{$12.9308_{(1)}$}   & \textcolor{green}{$0.1342_{(2)}$} & \textcolor{green}{$0.0995_{(2)}$} & 0.8554                          & \textcolor{blue}{$23.5434_{(3)}$}  & \textcolor{blue}{$24.3670_{(3)}$} \\
SuBSENSE                   & 19.5009                          & 0.2367                          & 0.1353                          & \textcolor{blue}{$0.8562_{(3)}$}  & 18.0343                & 19.2097 \\
PAWCS                      & 25.0482                          & 0.3244                          & 0.1752                          & 0.7800                          & 17.1190                & 18.2663 \\
TensorMoG                  & \textcolor{green}{$13.8699_{(2)}$} & \textcolor{blue}{$0.1446_{(3)}$}  & \textcolor{red}{$0.0745_{(1)}$}   & \textcolor{red}{$0.8769_{(1)}$}   & 22.1011                & 23.2473 \\
\textbf{CDN-GM}    & 19.6865                          & 0.2033                          & 0.1486                          & 0.8405                          & \textcolor{green}{$23.5831_{(2)}$} & \textcolor{green}{$24.5391_{(2)}$} \\
\bottomrule
\end{tabular}}{}
\resizebox{0.49\textwidth}{!}{\begin{tabular}{cccccc}      
\multicolumn{6}{p{10cm}}{\footnotesize In each column, \textcolor{red}{$Red_{(1)}$} is for the best, \textcolor{green}{$Green_{(2)}$} {is for} the second best, and~\textcolor{blue}{$Blue_{(3)}$} {is for} the third best.}
\end{tabular}}{}
\vspace{-5mm}
\end{table}

\subsection{Result on SBMnet benchmarks}
We perform an empirical ablation study of how the background generator, CDN-GM, contributes into the overall architecture with the SBMnet dataset \cite{Jodoin2017} for evaluating background estimation results. The SBMnet dataset has 80 real-life video sequences and their corresponding ground-truth backgrounds for references over 8 scenarios (illumination changes, cluttering, camera jitter, intermittent motion, etc.). 
Some of the algorithms that do not model the background, e.g., FgSegNet, SWCD, etc., are left out by default. For brevity, Table \ref{tab:sbmnet-comparison} provides the overall quantitative rankings (across all dataset sequences) of the proposed method along with state-of-the-art background estimation algorithms which are originally based on Gaussian Mixture Estimation.

\begin{table*}[b!]
\vspace{-3mm}
\caption{F - measure comparisons over all of ten categories in the CDnet 2014 dataset}
\label{tab:cdnet-fmeasure}
\centering
\resizebox{1.0\textwidth}{!}{
\begin{tabular}{c|lllllllllll|l}

\toprule
\multicolumn{1}{c|}{\textbf{~}} & \multicolumn{1}{c}{\textbf{Method}} & \multicolumn{1}{c}{\textit{\textbf{BDW}}} & \multicolumn{1}{c}{\textit{\textbf{LFR}}} & \multicolumn{1}{c}{\textit{\textbf{NVD}}} & \multicolumn{1}{c}{\textit{\textbf{THM}}} & \multicolumn{1}{c}{\textit{\textbf{SHD}}} & \multicolumn{1}{c}{\textit{\textbf{IOM}}} & \multicolumn{1}{c}{\textit{\textbf{CJT}}} & \multicolumn{1}{c}{\textit{\textbf{DBG}}} & \multicolumn{1}{c}{\textit{\textbf{BSL}}} & \multicolumn{1}{c|}{\textit{\textbf{TBL}}} & \multicolumn{1}{c}{\textit{\textbf{Average}}} \\ 

\midrule
\parbox[t]{2mm}{\multirow{8}{*}{\rotatebox[origin=c]{90}{Unsupervised}}}
                              & GMM -- S \& G  & 0.7380                   & 0.5373                   & 0.4097                   & 0.6621                   & 0.7156                   & 0.5207                   & 0.5969                   & 0.6330                   & 0.8245                   & 0.4663                  & 0.6104 \\ 
                              & GMM -- Zivkovic             & 0.7406                   & 0.5065                   & 0.3960                   & 0.6548                   & 0.7232                   & 0.5325                   & 0.5670                   & 0.6328                   & 0.8382                   & 0.4169                  & 0.6009 \\ 
                              & SuBSENSE                    & \textcolor{green}{$0.8619_{(2)}$}                   & 0.6445                   & \textcolor{blue}{$0.5599_{(3)}$}                   & \textcolor{blue}{$0.8171_{(3)}$}                   & \textcolor{blue}{$0.8646_{(3)}$}                   & 0.6569                   & \textcolor{green}{$0.8152_{(2)}$}                   & 0.8177                   & \textcolor{red}{$0.9503_{(1)}$}                   & \textcolor{green}{$0.7792_{(2)}$}                  & 0.7767 \\ 
                              & PAWCS                       & 0.8152                   & \textcolor{blue}{$0.6588_{(3)}$}                   & 0.4152                   & \textcolor{red}{$0.9921_{(1)}$}                   & \textcolor{green}{$0.8710_{(2)}$}                   & \textcolor{blue}{$0.7764                  _{(3)}$} & \textcolor{blue}{$0.8137_{(3)}$}                   & \textcolor{red}{$0.8938_{(1)}$}                   & \textcolor{blue}{$0.9397_{(3)}$}                   & 0.6450                   & \textcolor{blue}{$0.7821_{(3)}$}                  \\ 
                              & TensorMoG                   & \textcolor{red}{$0.9298_{(1)}$}                   & \textcolor{green}{$0.6852_{(2)}$}                   & \textcolor{green}{$0.5604_{(2)}$}                   & 0.7993                   & \textcolor{red}{$0.9738_{(1)}$}                   & \textcolor{red}{$0.9325_{(1)}$}                   & \textcolor{red}{$0.9325_{(1)}$}                   & 0.6493                   & \textcolor{green}{$0.9488_{(2)}$}                   & \textcolor{red}{$0.8380_{(1)}$}                  & \textcolor{red}{$0.7977_{(1)}$}                  \\ 
                              & BMOG                        & 0.7836                   & 0.6102                   & 0.4982                   & 0.6348                   & 0.8396                   & 0.5291                   & 0.7493                   & 0.7928                   & 0.8301                   & 0.6932                  & 0.6961 \\ 
                              & FTSG                        & 0.8228                   & 0.6259                   & 0.5130                   & 0.7768                   & 0.8535                   & \textcolor{green}{$0.7891_{(2)}$}                   & 0.7513                   & \textcolor{green}{$0.8792_{(2)}$}                   & 0.9330                   & 0.7127                  & 0.7657 \\ 
                              & SWCD                        & \textcolor{blue}{$0.8233_{(3)}$}                   & \textcolor{red}{$0.7374_{(1)}$}                   & \textcolor{red}{$0.5807_{(1)}$}                   & \textcolor{green}{$0.8581_{(2)}$}                   & 0.8302                   & 0.7092                   & 0.7411                   & \textcolor{blue}{$0.8645_{(3)}$}                   & 0.9214                   & \textcolor{blue}{$0.7735_{(3)}$}                  & \textcolor{green}{$0.7839_{(2)}$} \\ 
\midrule
\parbox[t]{2mm}{\multirow{1}{*}{\rotatebox[origin=c]{90}{$\ast$}}}
& \textbf{CDN-MEDAL-net}    & \textbf{0.9045}                   & \textbf{0.9561}                   & \textbf{0.8450}                   & \textbf{0.9129}                   & \textbf{0.8683}                   & \textbf{0.8249}                   & \textbf{0.8427}                   & \textbf{0.9372}                   & \textbf{0.9615}                   & \textbf{0.9187}                   &\textbf{0.8972} \\ 
\midrule
\parbox[t]{2mm}{\multirow{6}{*}{\rotatebox[origin=c]{90}{Supervised}}}
& FgSegNet\_S    & \textcolor{green}{$0.9897_{(2)}$}                   & \textcolor{green}{$0.8972_{(2)}$}                   & \textcolor{green}{$0.9713_{(2)}$}                   & \textcolor{red}{$0.9921 _{(1)}$}                  & \textcolor{blue}{$0.9937_{(3)}$}                   & \textcolor{blue}{$0.9940_{(3)}$}                    & \textcolor{green}{$0.9957_{(2)}$}                   & \textcolor{green}{$0.9958_{(2)}$}                   & \textcolor{red}{$0.9977_{(1)}$}                   & 0.9681            & \textcolor{green}{$0.9795_{(2)}$}       
\\ 
& FgSegNet                    & \textcolor{blue}{$0.9845_{(3)}$}                   & \textcolor{blue}{$0.8786_{(3)}$}                   & \textcolor{blue}{$0.9655_{(3)}$}                   & \textcolor{blue}{$0.9648_{(3)}$}                   & \textcolor{green}{$0.9973_{(2)}$}                   & \textcolor{red}{$0.9958_{(1)}$}                   & \textcolor{blue}{$0.9954_{(3)}$}                   & \textcolor{blue}{$0.9951_{(3)}$}                   & \textcolor{blue}{$0.9944_{(3)}$}                   & \textcolor{green}{$0.9921_{(2)}$}                  & \textcolor{blue}{$0.9764_{(3)}$} \\ 
& FgSegNet\_v2                & \textcolor{red}{$0.9904_{(1)}$}                   & \textcolor{red}{$0.9336_{(1)}$}                   & \textcolor{red}{$0.9739_{(1)}$}                   & \textcolor{green}{$0.9727_{(2)}$}                   & \textcolor{red}{$0.9978_{(1)}$}                   & \textcolor{green}{$0.9951_{(2)}$}                   & \textcolor{red}{$0.9971_{(1)}$}                   & \textcolor{red}{$0.9961_{(1)}$}                   & \textcolor{green}{$0.9952_{(2)}$}                   & \textcolor{red}{$0.9938_{(1)}$}                  & \textcolor{red}{$0.9846_{(1)}$}                  \\ 
& Cascade CNN          & 0.9431                   & 0.8370                   & 0.8965                   & 0.8958                   & 0.9414                   & 0.8505                   & 0.9758                   & 0.9658                   & 0.9786                   & 0.9108                   & 0.9195 \\  
& DeepBS               & 0.8301                   & 0.6002                   & 0.5835                   & 0.7583                   & 0.9092                   & 0.6098                   & 0.8990                   & 0.8761                   & 0.9580                   & 0.8455                   & 0.7870 \\ 
& STAM                        & 0.9703                   & 0.6683                   & 0.7102                   & 0.9328                   & 0.9885                   & 0.9483                   & 0.8989                   & 0.9155                   & 0.9663                   & \textcolor{blue}{$0.9907_{(3)}$}                  & 0.8989 \\ 
\bottomrule
\end{tabular}}
\resizebox{1.0\textwidth}{!}{\begin{tabular}{cccccc}      
\multicolumn{6}{p{20cm}}{\footnotesize ${}^{\ast}$Semi-Unsupervised; Experimented scenarios include bad weather (\textit{BDW}), low frame rate (\textit{LFR}), night videos (\textit{NVD}), turbulence (\textit{TBL}), baseline (\textit{BSL}), dynamic background (\textit{DBG}), camera jitter (\textit{CJT}), intermittent object motion (\textit{IOM}), shadow (\textit{SHD}), and thermal (\textit{THM}). In each column, \textcolor{red}{$Red_{(1)}$} is for the best, \textcolor{green}{$Green_{(2)}$} {is for} the second best, and~\textcolor{blue}{$Blue_{(3)}$} {is for} the third best.}
\end{tabular}}
\vspace{-5mm}
\end{table*}

In general, traditional GMM-based methods like GMM - Stauffer \& Grimson, GMM - Zivkovic, and TensorMoG, are the top-performing methods in the background modeling domain. The proposed CDN-GM module is outperformed by these traditional GMM-based algorithms in terms of the pixel-based metrics (i.e., AGE, pEPs, and pCEPs), which measure the intensity difference between the generated background and the ground-truth. However, the gains of CDN-GM on visual quality measurements (i.e., MS-SSIM, PSNR, and CQM) signify that the background generated by CDN-GM is competitive against the top GMM-based methods on the background estimation domain in terms of textural and semantic information compared to the ground-truth.

The shortcomings of the proposed background extraction methods in its exact background grayscale estimation shows up very clearly in the three metrics AGE, pEPs, and pCEPs, where lower results are better. 
There are two main possible reasons why such shortcomings exist. First, the first three grayscale-based metrics are highly sensitive to small variations in the estimated background as these metrics measure the estimation result based entirely on its absolute difference with the provided ground-truth. In real life, however, 
avoiding variations in capturing pixel signals is inherently impossible. Thus, while these metrics surely provide some degree of confidence in the computed background image quality, they cannot serve as the absolute determination of the image quality. Second, our design of CDN-GM focuses on speed efficiency with a small temporal window in contrast to traditional GMM-based methods that can capture long-term pixel signals to estimate the background intensities with high accuracy. This trade-off between the efficiency and effectiveness results in a clear disadvantage for CDN-GM in estimating the grayscale signal as close as possible to the groundtruth. However, the absolute difference of the CDN-GM module with the other five methods on each metric is still within an acceptable margin: 6.7557 in AGE (compared to GMM - Zivkovic), 0.0758 in pEPs (compared to GMM - Stauffer \& Grimson), and 0.0741 in pCEPs (compared to TensorMoG).

In contrast, the other three metrics (MS-SSIM, PSNR, and CQM) measure the visual and structural distortion of estimated backgrounds against groundtruths. The focus of these metrics was to quantify the errors in artificially generated image's visual quality as perceived by humans as closely as possible. In these metrics, higher quantitative values correspond to better background estimations. Our CDN-GM's background images' quality is showcased via (1) the difference between the top-1 method in MS-SSIM and TensorMoG with parameter tuning, is only a marginal value of 0.0364, and (2) CDN-GM consistently shows up as the second best method in PSNR and CQM. Thus, backgrounds approximated by CDN-GM are images of good textural quality compared to ground-truths.

As an unsupervised, generalistic approach, 
the results on visual quality metrics implies that the semantic information of the generated background by our proposed CDN-GM and the input image is very similar. This semantic similarity between the background and the input frame has suppressed a large amount of background distractors from the input frame for the imbalance foreground segmentation learning task (very high difference in the number of pixels classified as background compared to the number of foreground pixels). 
Thus, CDN-GM is advantageous in maintaining adaptation in cases that environmental changes happen often (e.g., illumination changes) like traditional GMM approaches, its learning to generalize for a small local temporal history is comparable to slow, gradual adaptation of GMM-based of algorithms.


\subsection{Results on CDnet 2014 Benchmarks}
With 53 video sequences (length varying from 1000 to 7000 frames) spread over 11 different scenarios, the CDnet-2014 dataset \cite{Wang2014} is the current biggest, most comprehensive large-scale public dataset for evaluating algorithms in change detection. We demonstrate empirically the effectiveness of our proposed approach across a plethora of scenarios and effects. 

Our proposed method was trained on CDnet-2014 dataset for about 1000 epochs for each sequence in CDnet using Adam optimizer with the learning rate = $5e^{-3}$. The training dataset for MEDAL-net is sampled so that 200 labeled foregrounds per video sequence are chosen to train the model. This is only up to 20\% of the number of labeled frames for some sequences in CDnet, and 8.7\% of CDnet's labeled data in overall. During training, the associated background of each chosen frame is directly generated using CDN-GM as MEDAL-net is trained separately from CDN-GM because of the manually chosen input-label pairs. This strategy of sampling for supervised learning is the same as that of FgSegNet's and Cascade CNN. 

\begin{figure*}[!t]
\vspace{-8mm}
\setlength\tabcolsep{1.5pt} 
\renewcommand{\arraystretch}{2.7}
    \resizebox{0.83\textwidth}{!}{
    \begin{tabular*}{1.0\textwidth}{@{}rcccccccccccccccc@{}}
    
    &   
    ($\star$) &
    ($\diamond$) &
    (a) &
    (b) &
    (c) &
    (d) &
    (e) & 
    (f) &
    (g) &
    (h) &
    (i) &
    (j) &
    (k) &
    (l)
    \\
    
    \arrayrulecolor{red}\cline{11-11}    
    \rule{0pt}{25pt}\parbox[t]{2mm}{\rotatebox[origin=c]{90}{BDW}}  &
    \includegraphics[align=c, width=0.08\textwidth, height = 37pt]{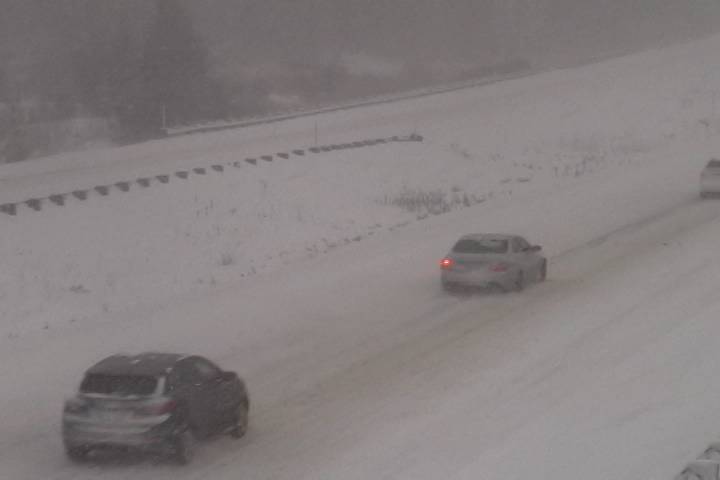} &
    \includegraphics[align=c, width=0.08\textwidth, height = 37pt]{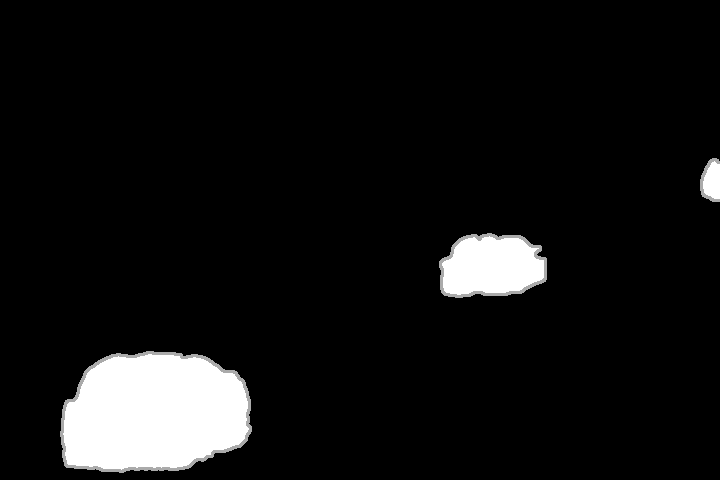} &
    \includegraphics[align=c, width=0.08\textwidth, height = 37pt]{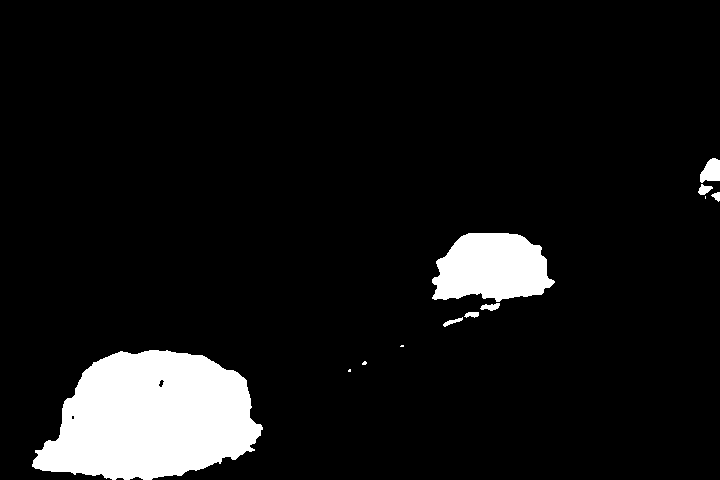} & 
    \includegraphics[align=c, width=0.08\textwidth, height = 37pt]{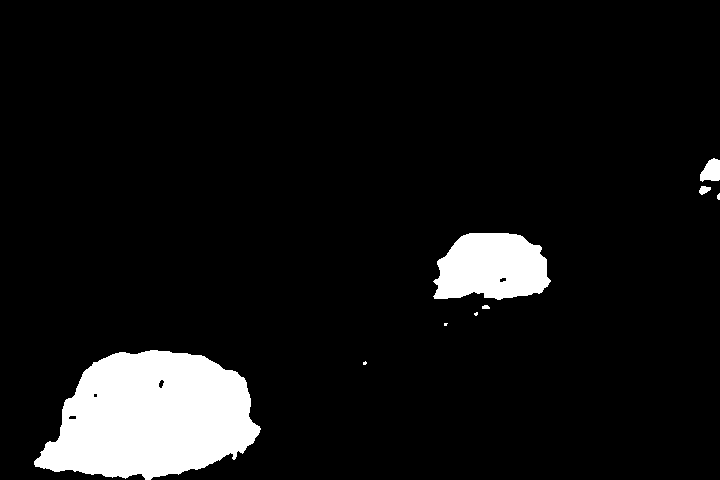} & 
    \includegraphics[align=c, width=0.08\textwidth, height = 37pt]{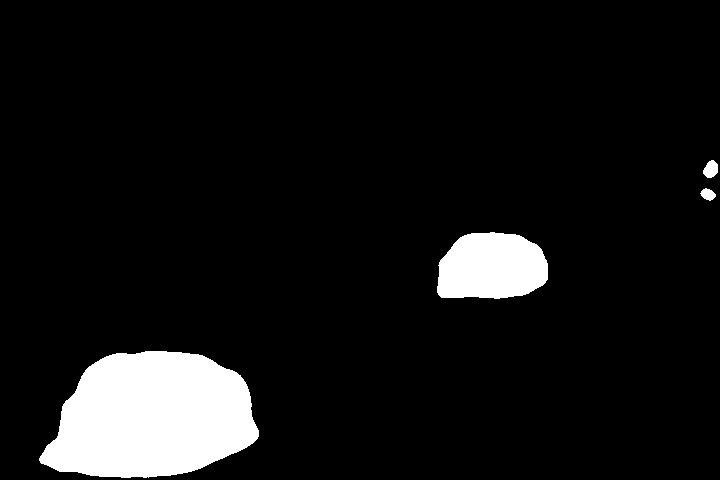} &
    \includegraphics[align=c, width=0.08\textwidth, height = 37pt]{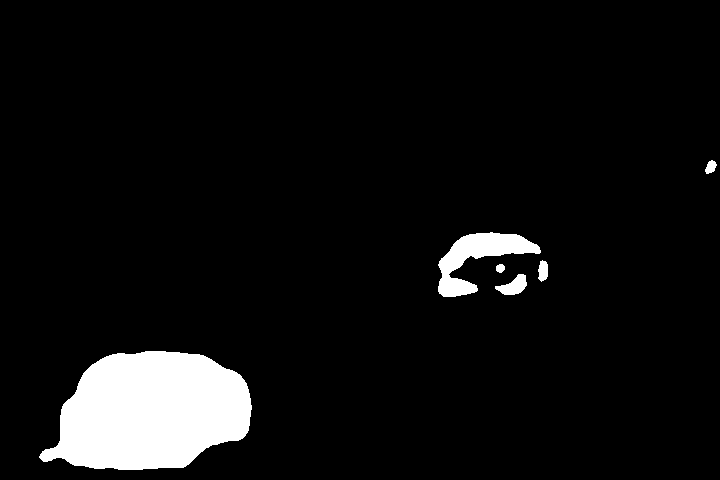} &
    \includegraphics[align=c, width=0.08\textwidth, height = 37pt]{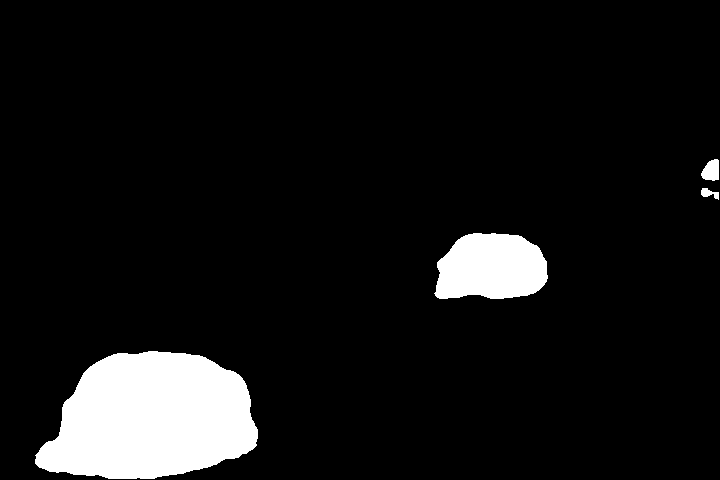} &
    \includegraphics[align=c, width=0.08\textwidth, height = 37pt]{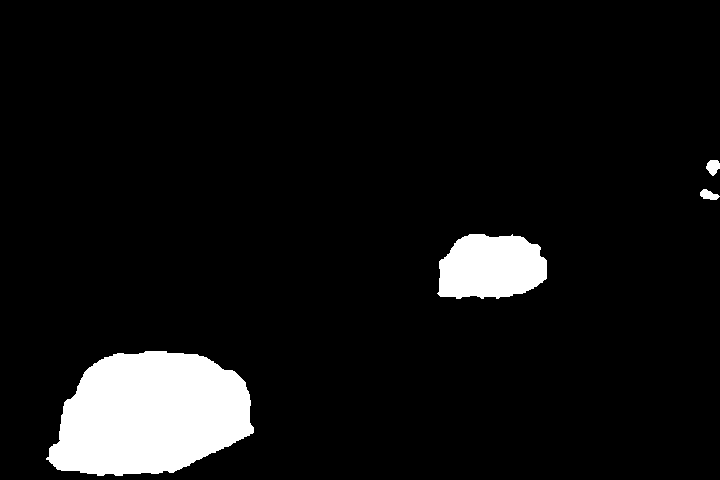} &
    \includegraphics[align=c, width=0.08\textwidth, height = 37pt]{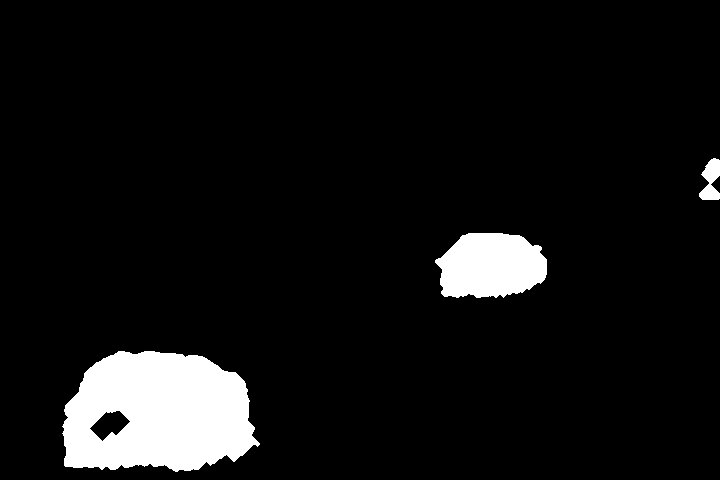} &     
    \Thickvrule{\includegraphics[align=c, width=0.08\textwidth, height = 37pt]{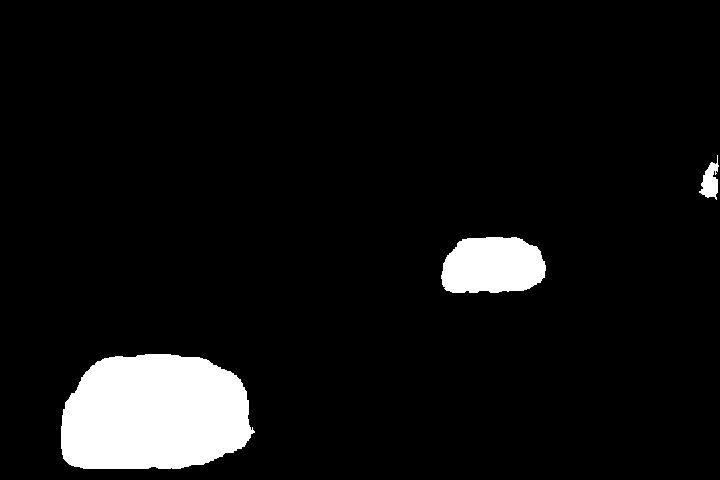}} &
    \includegraphics[align=c, width=0.08\textwidth, height = 37pt]{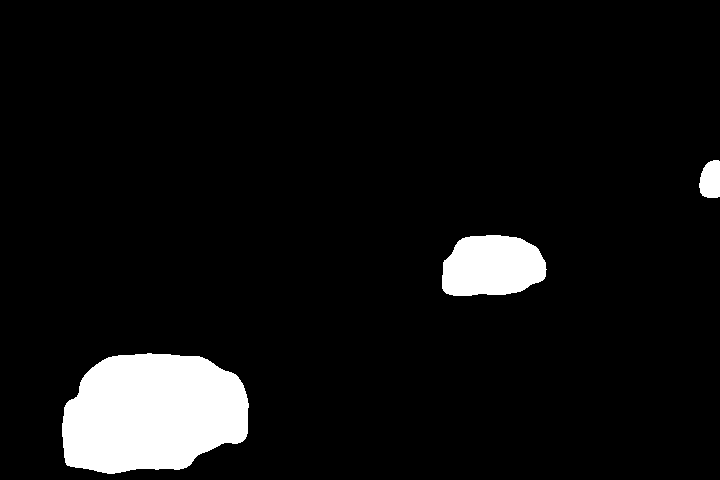} &
    \includegraphics[align=c, width=0.08\textwidth, height = 37pt]{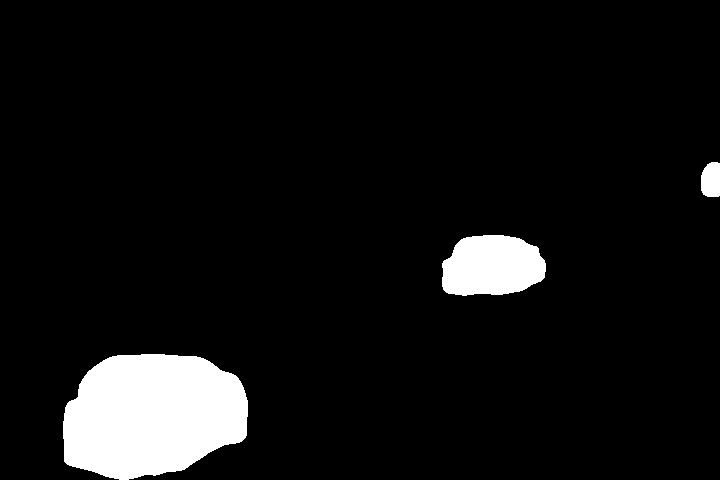} &
    \includegraphics[align=c, width=0.08\textwidth, height = 37pt]{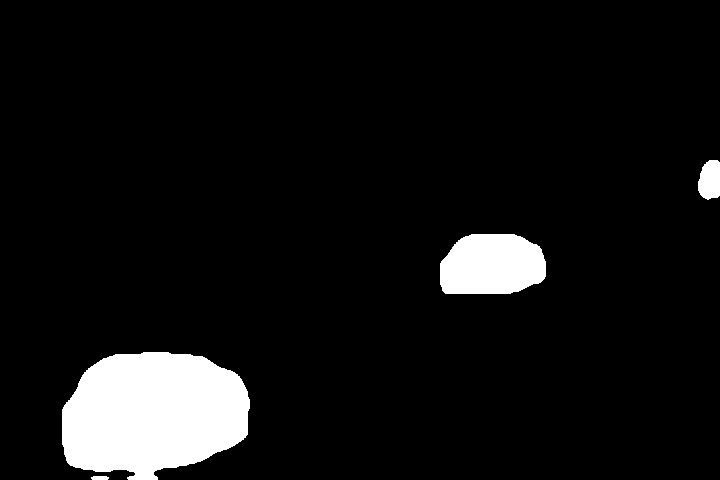} &
    \includegraphics[align=c, width=0.08\textwidth, height = 37pt]{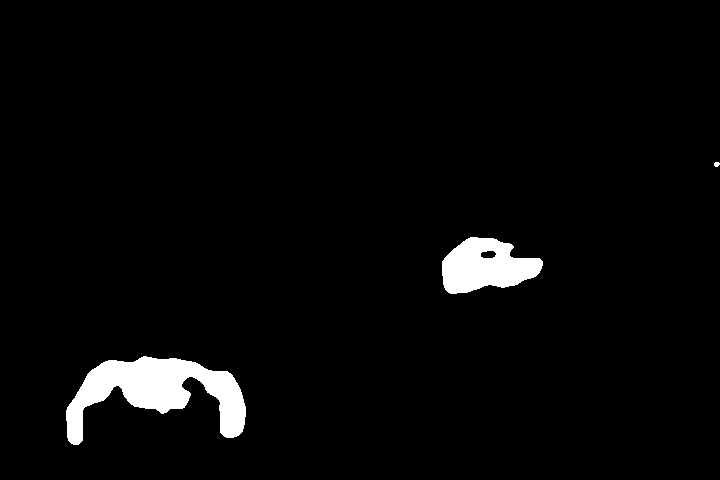} &
    \\
                             
    \parbox[t]{2mm}{\rotatebox[origin=c]{90}{BSL}}  &
    \includegraphics[align=c, width=0.08\textwidth, height = 37pt]{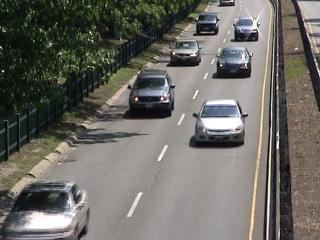} &
    \includegraphics[align=c, width=0.08\textwidth, height = 37pt]{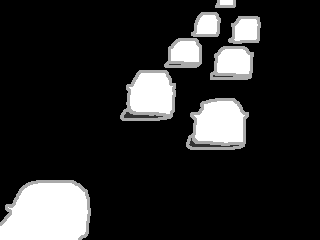} &
    \includegraphics[align=c, width=0.08\textwidth, height = 37pt]{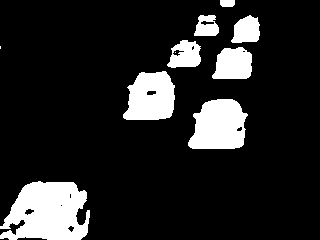} & 
    \includegraphics[align=c, width=0.08\textwidth, height = 37pt]{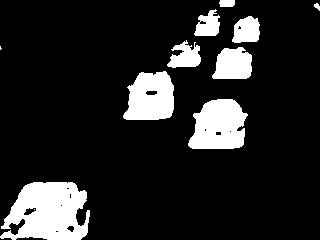} & 
    \includegraphics[align=c, width=0.08\textwidth, height = 37pt]{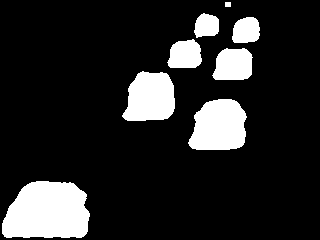} &
    \includegraphics[align=c, width=0.08\textwidth, height = 37pt]{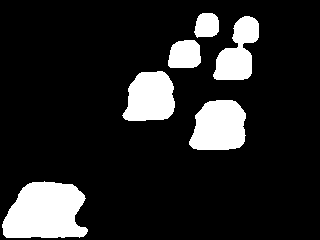} &
    \includegraphics[align=c, width=0.08\textwidth, height = 37pt]{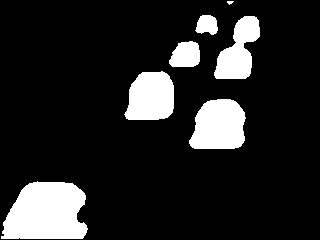} &
    \includegraphics[align=c, width=0.08\textwidth, height = 37pt]{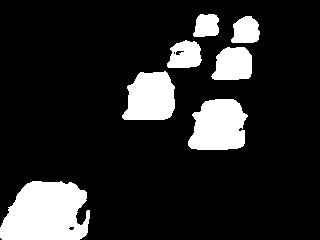} &
    \includegraphics[align=c, width=0.08\textwidth, height = 37pt]{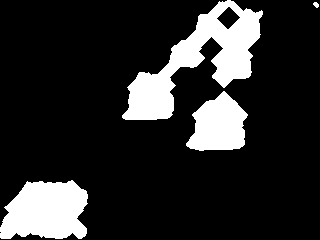} &     
    \Thickvrule{\includegraphics[align=c, width=0.08\textwidth, height = 37pt]{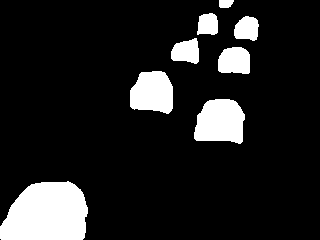}} &
    \includegraphics[align=c, width=0.08\textwidth, height = 37pt]{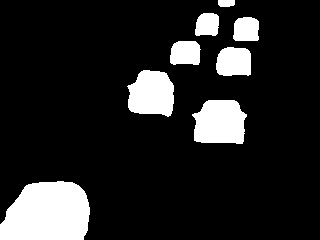} &
    \includegraphics[align=c, width=0.08\textwidth, height = 37pt]{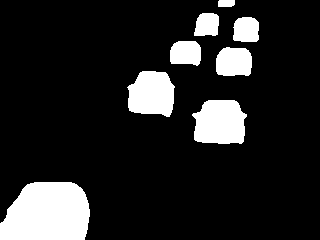} &
    \includegraphics[align=c, width=0.08\textwidth, height = 37pt]{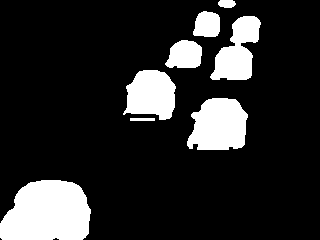} &
    \includegraphics[align=c, width=0.08\textwidth, height = 37pt]{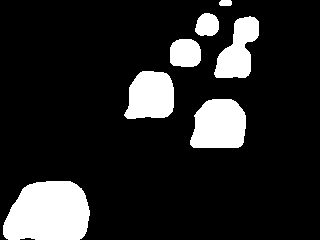} &
    \\
    
    \parbox[t]{2mm}{\rotatebox[origin=c]{90}{CJT}}  &
    \includegraphics[align=c, width=0.08\textwidth, height = 37pt]{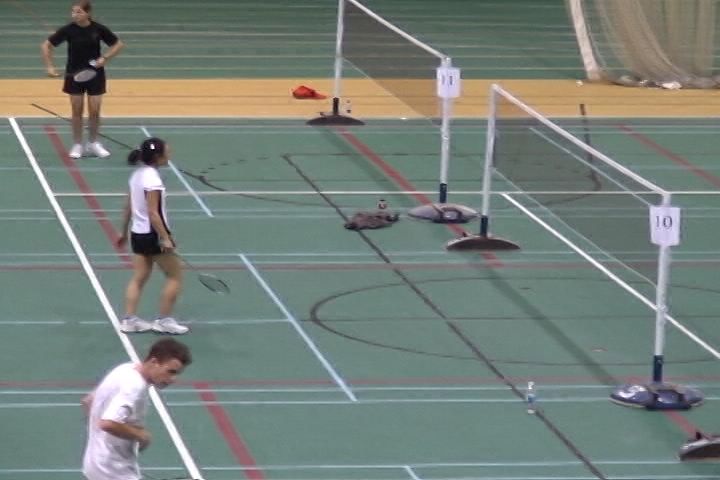} &
    \includegraphics[align=c, width=0.08\textwidth, height = 37pt]{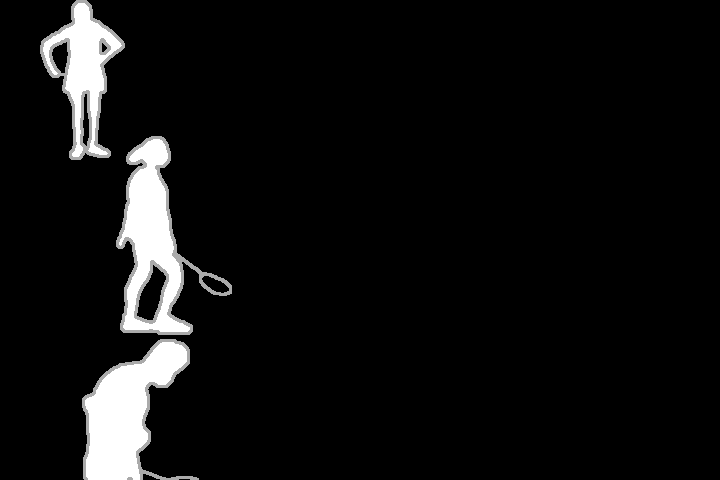} &
    \includegraphics[align=c, width=0.08\textwidth, height = 37pt]{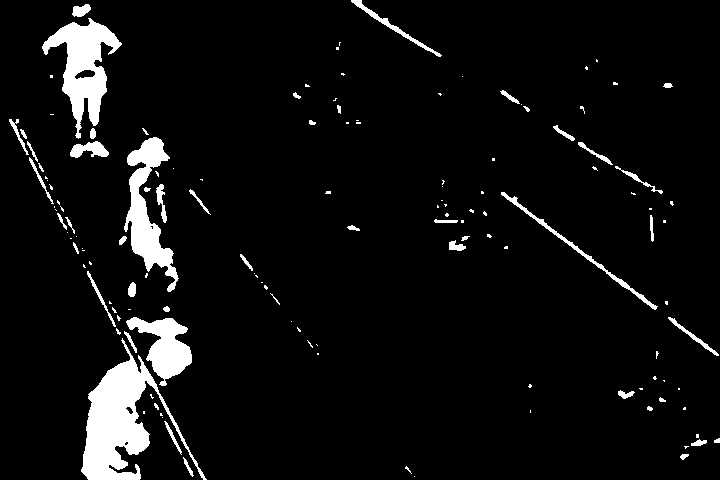} & 
    \includegraphics[align=c, width=0.08\textwidth, height = 37pt]{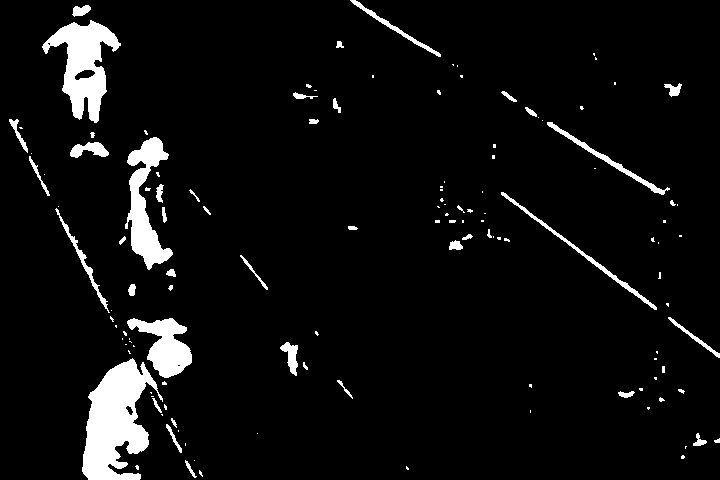} & 
    \includegraphics[align=c, width=0.08\textwidth, height = 37pt]{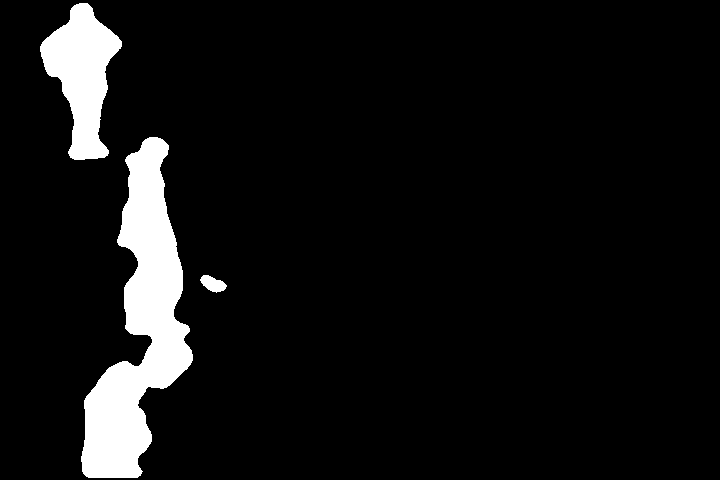} &
    \includegraphics[align=c, width=0.08\textwidth, height = 37pt]{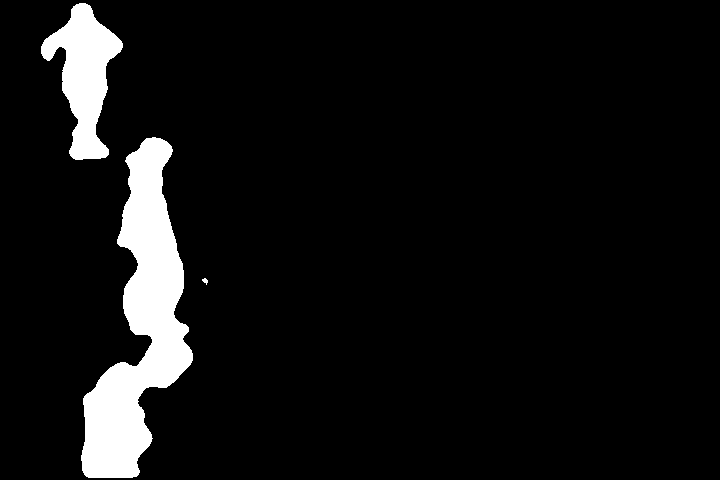} &
    \includegraphics[align=c, width=0.08\textwidth, height = 37pt]{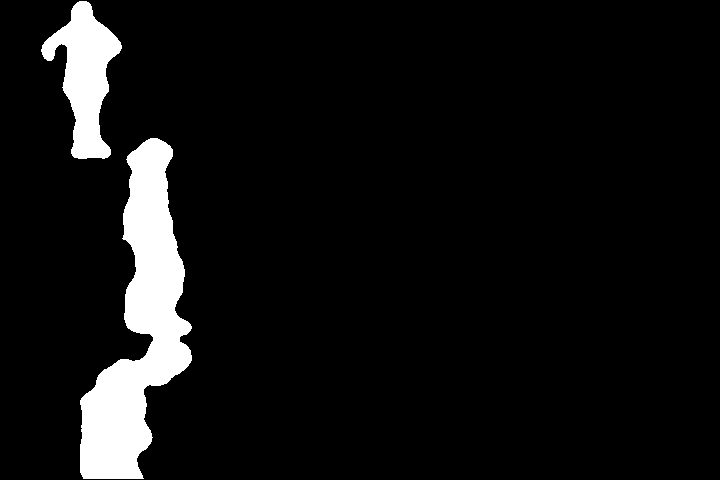} &
    \includegraphics[align=c, width=0.08\textwidth, height = 37pt]{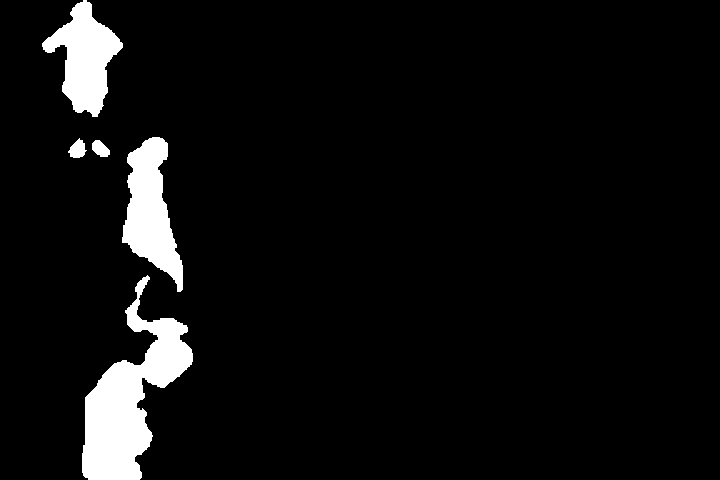} &
    \includegraphics[align=c, width=0.08\textwidth, height = 37pt]{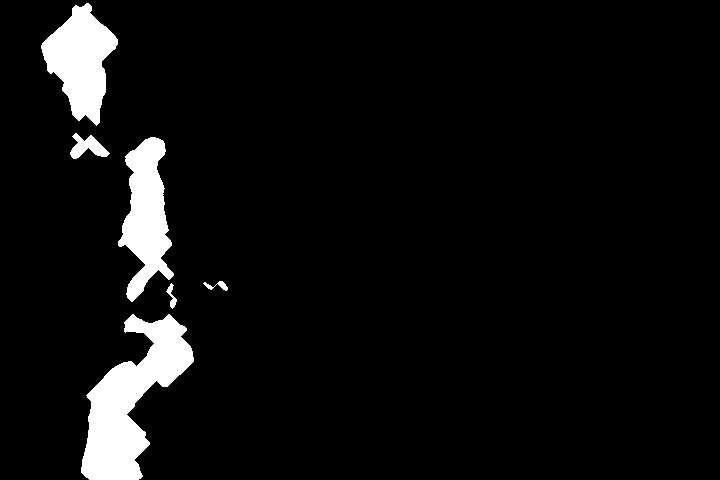} &     
    \Thickvrule{\includegraphics[align=c, width=0.08\textwidth, height = 37pt]{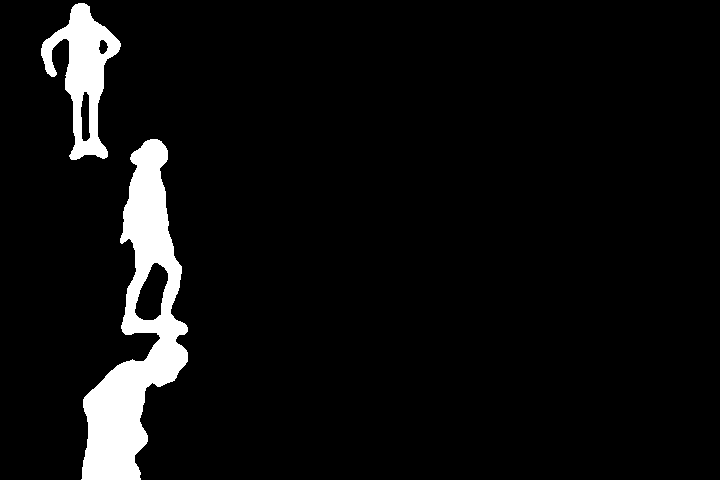}} &
    \includegraphics[align=c, width=0.08\textwidth, height = 37pt]{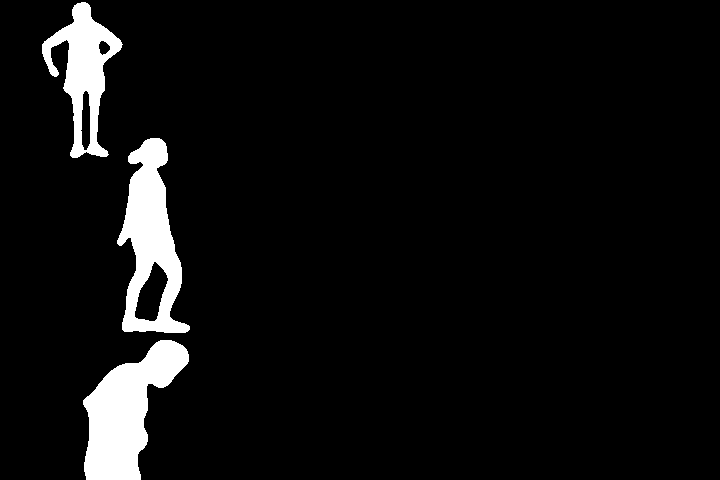} &
    \includegraphics[align=c, width=0.08\textwidth, height = 37pt]{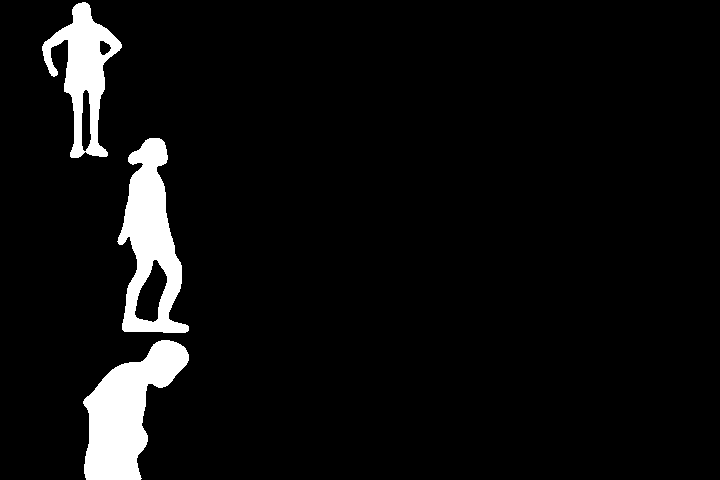} &
    \includegraphics[align=c, width=0.08\textwidth, height = 37pt]{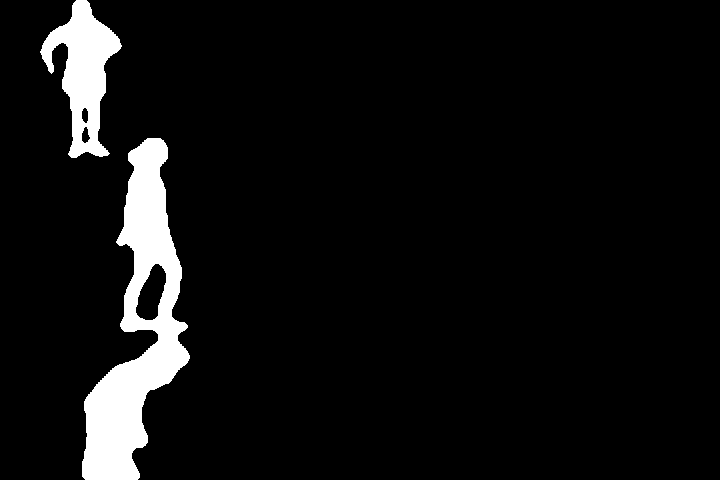} &
    \includegraphics[align=c, width=0.08\textwidth, height = 37pt]{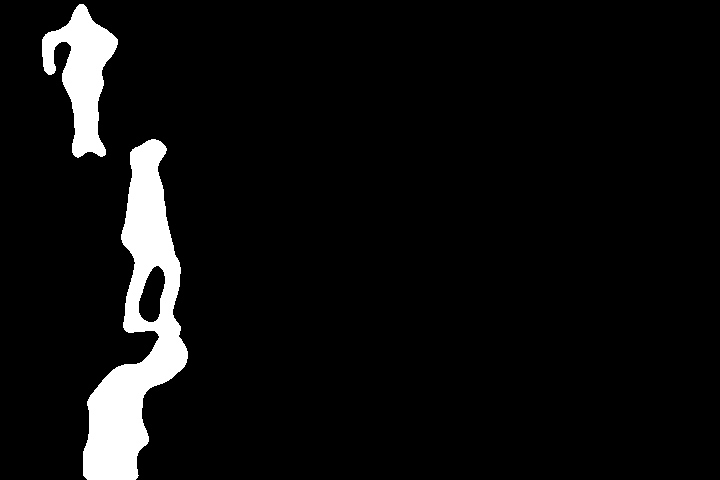} &
    \\
                           
    \parbox[t]{2mm}{\rotatebox[origin=c]{90}{DBG}}  &
    \includegraphics[align=c, width=0.08\textwidth, height = 37pt]{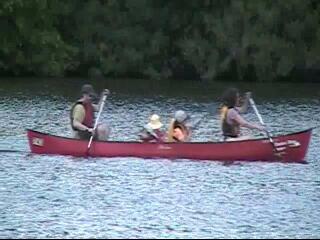} &
    \includegraphics[align=c, width=0.08\textwidth, height = 37pt]{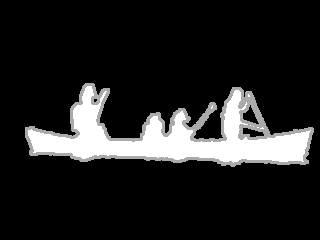} &
    \includegraphics[align=c, width=0.08\textwidth, height = 37pt]{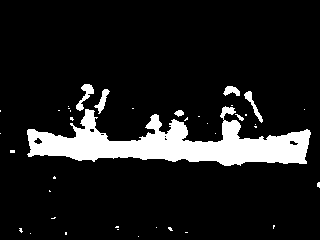} & 
    \includegraphics[align=c, width=0.08\textwidth, height = 37pt]{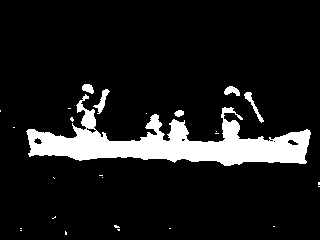} & 
    \includegraphics[align=c, width=0.08\textwidth, height = 37pt]{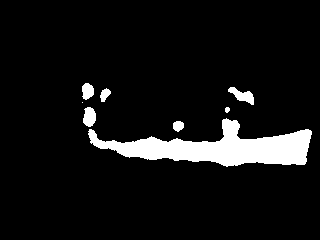} &
    \includegraphics[align=c, width=0.08\textwidth, height = 37pt]{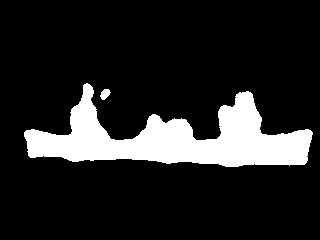} &
    \includegraphics[align=c, width=0.08\textwidth, height = 37pt]{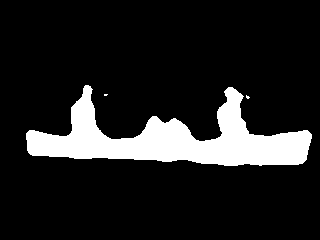} &
    \includegraphics[align=c, width=0.08\textwidth, height = 37pt]{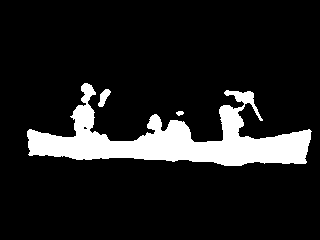} &
    \includegraphics[align=c, width=0.08\textwidth, height = 37pt]{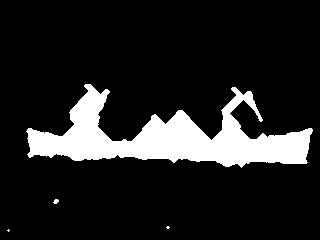} & 
    \Thickvrule{\includegraphics[align=c, width=0.08\textwidth, height = 37pt]{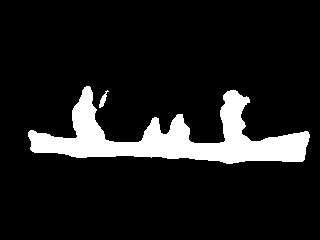}} &
    \includegraphics[align=c, width=0.08\textwidth, height = 37pt]{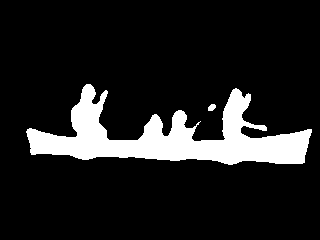} &
    \includegraphics[align=c, width=0.08\textwidth, height = 37pt]{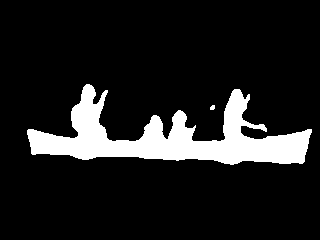} &
    \includegraphics[align=c, width=0.08\textwidth, height = 37pt]{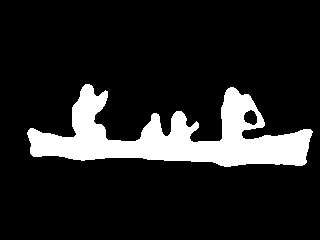} &
    \includegraphics[align=c, width=0.08\textwidth, height = 37pt]{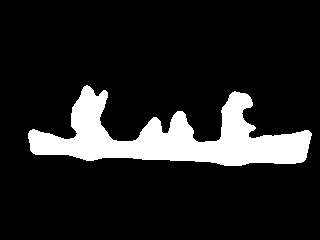} &
    \\
                           
    \parbox[t]{2mm}{\rotatebox[origin=c]{90}{IOM}}  &
    \includegraphics[align=c, width=0.08\textwidth, height = 37pt]{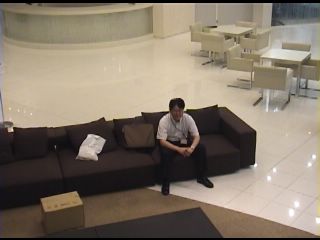} &
    \includegraphics[align=c, width=0.08\textwidth, height = 37pt]{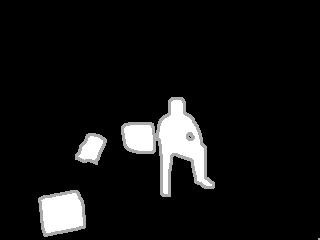} &
    \includegraphics[align=c, width=0.08\textwidth, height = 37pt]{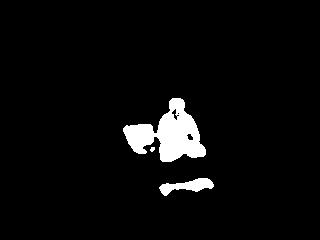} & 
    \includegraphics[align=c, width=0.08\textwidth, height = 37pt]{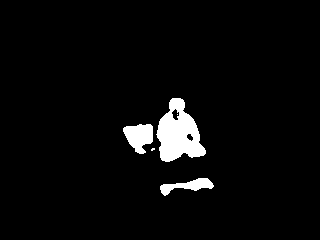} & 
    \includegraphics[align=c, width=0.08\textwidth, height = 37pt]{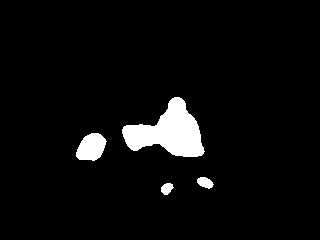} &
    \includegraphics[align=c, width=0.08\textwidth, height = 37pt]{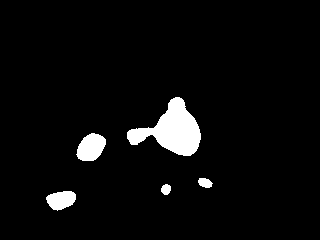} &
    \includegraphics[align=c, width=0.08\textwidth, height = 37pt]{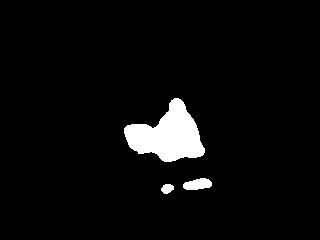} &
    \includegraphics[align=c, width=0.08\textwidth, height = 37pt]{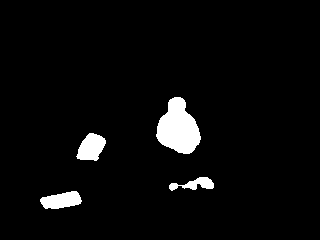} &
    \includegraphics[align=c, width=0.08\textwidth, height = 37pt]{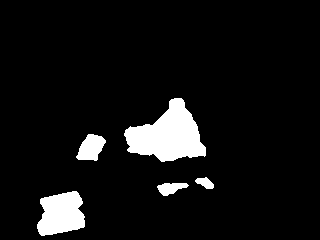} &  
    \Thickvrule{\includegraphics[align=c, width=0.08\textwidth, height = 37pt]{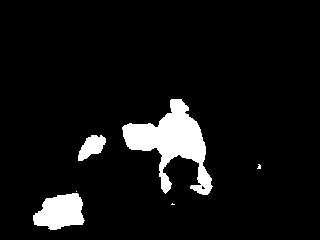}} &
    \includegraphics[align=c, width=0.08\textwidth, height = 37pt]{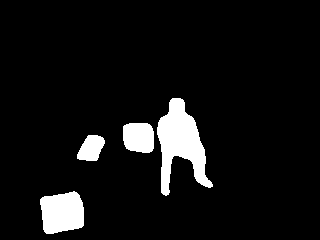} &
    \includegraphics[align=c, width=0.08\textwidth, height = 37pt]{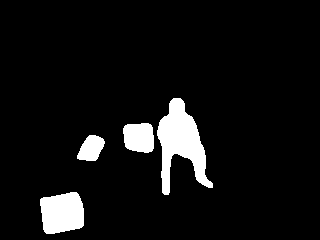} &
    \includegraphics[align=c, width=0.08\textwidth, height = 37pt]{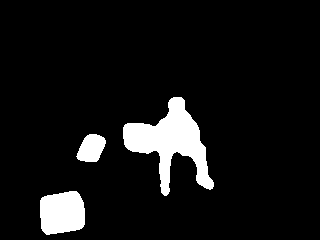} &
    \includegraphics[align=c, width=0.08\textwidth, height = 37pt]{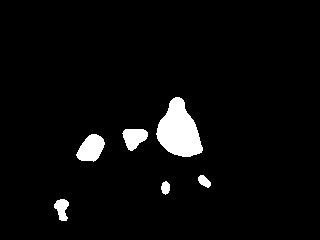} &
    \\
          
    \parbox[t]{2mm}{\rotatebox[origin=c]{90}{LFR}}  &
    \includegraphics[align=c, width=0.08\textwidth, height = 37pt]{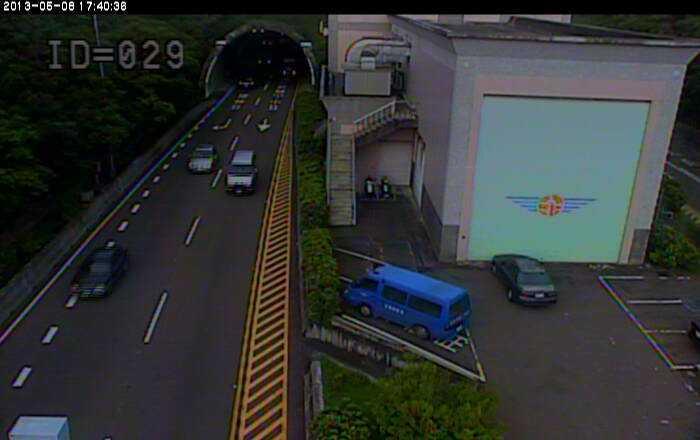} &
    \includegraphics[align=c, width=0.08\textwidth, height = 37pt]{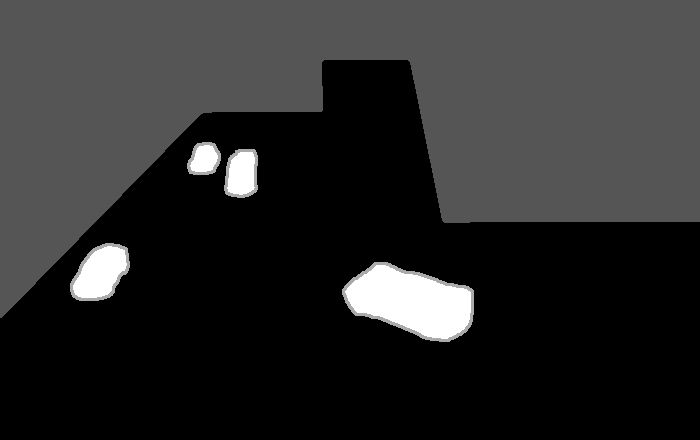} &
    \includegraphics[align=c, width=0.08\textwidth, height = 37pt]{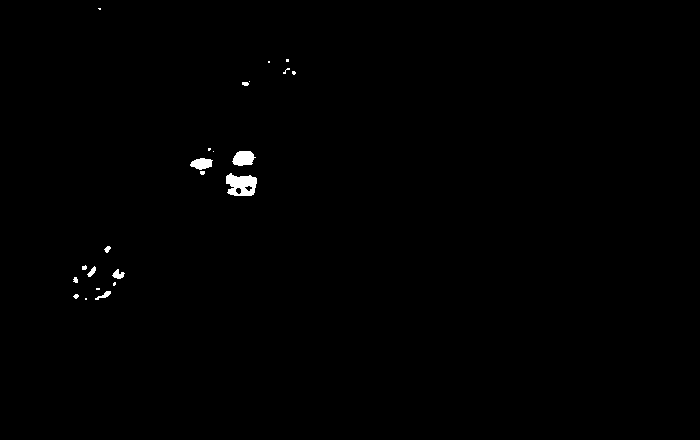} & 
    \includegraphics[align=c, width=0.08\textwidth, height = 37pt]{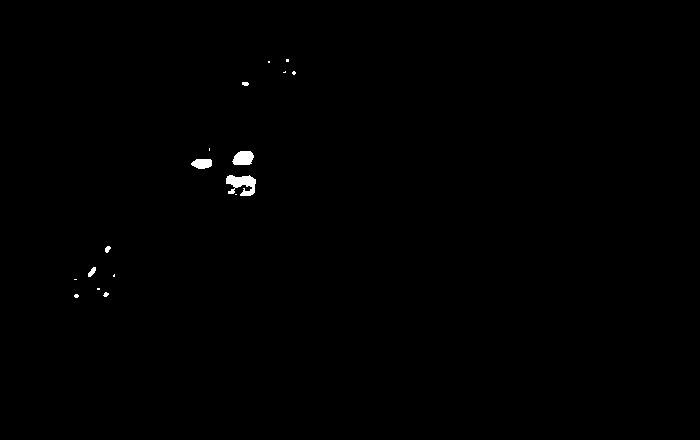} & 
    \includegraphics[align=c, width=0.08\textwidth, height = 37pt]{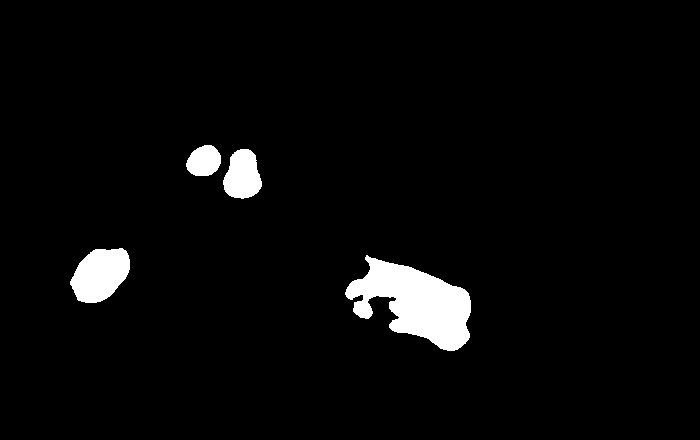} &
    \includegraphics[align=c, width=0.08\textwidth, height = 37pt]{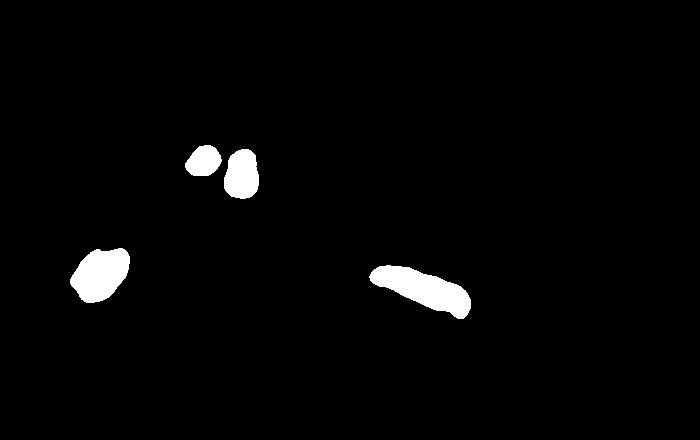} &
    \includegraphics[align=c, width=0.08\textwidth, height = 37pt]{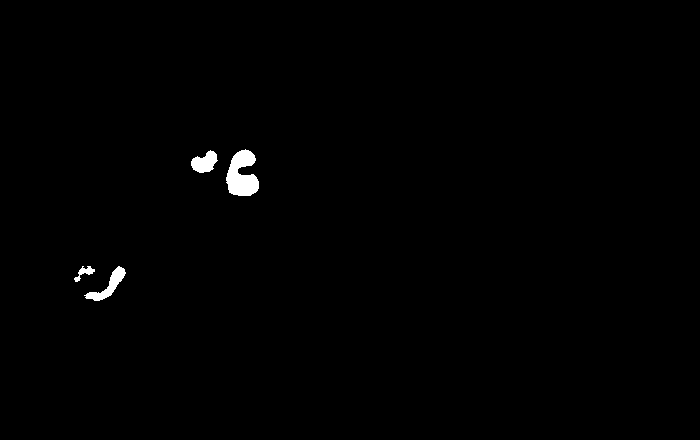} &
    \includegraphics[align=c, width=0.08\textwidth, height = 37pt]{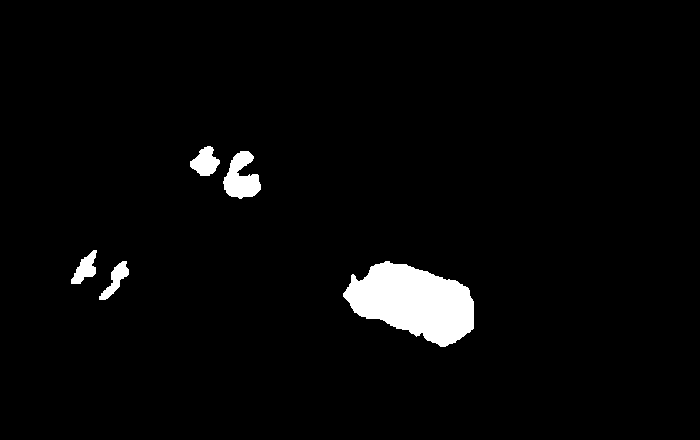} &
    \includegraphics[align=c, width=0.08\textwidth, height = 37pt]{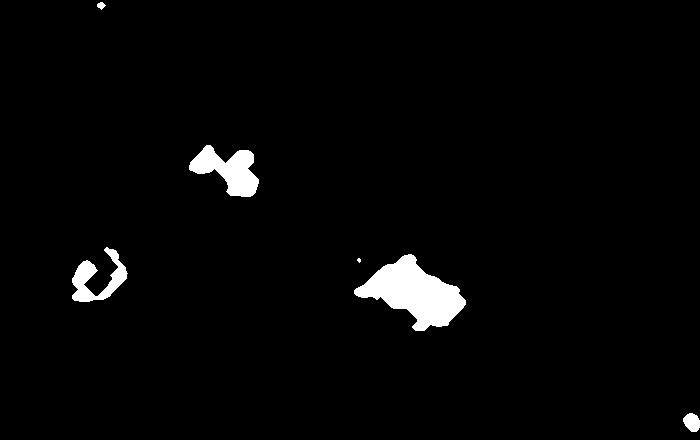} & 
    \Thickvrule{\includegraphics[align=c, width=0.08\textwidth, height = 37pt]{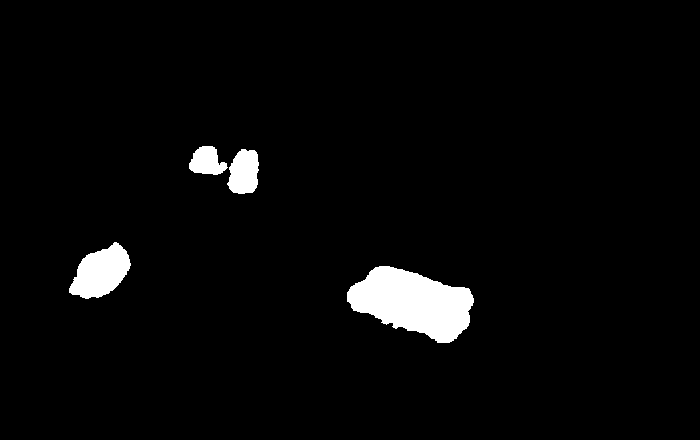}} &
    \includegraphics[align=c, width=0.08\textwidth, height = 37pt]{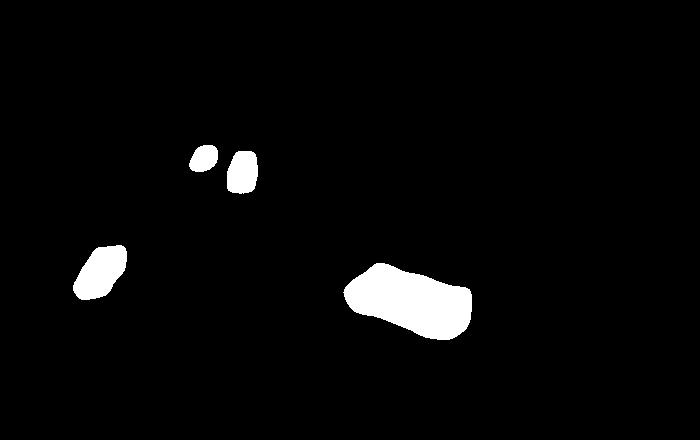} &
    \includegraphics[align=c, width=0.08\textwidth, height = 37pt]{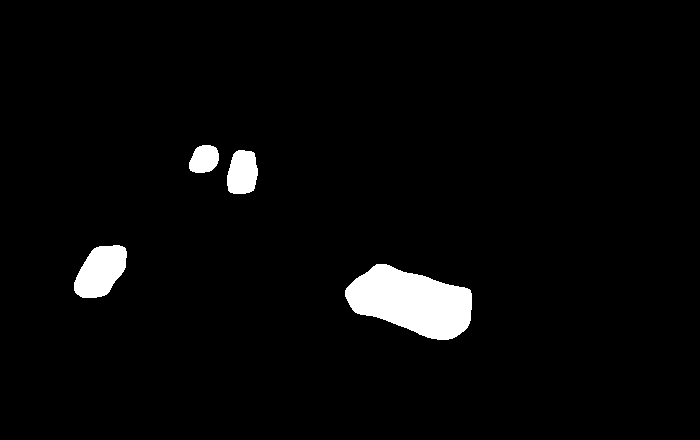} &
    \includegraphics[align=c, width=0.08\textwidth, height = 37pt]{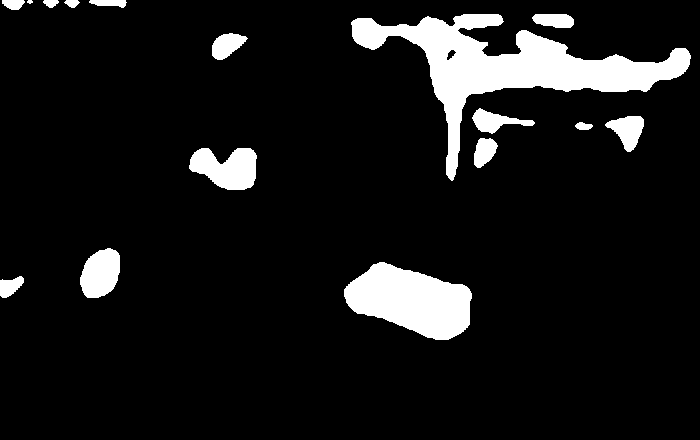} &
    \includegraphics[align=c, width=0.08\textwidth, height = 37pt]{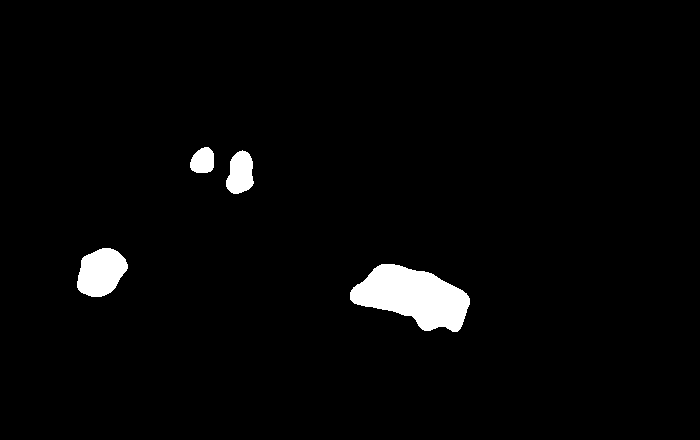} &
    \\
    
    \parbox[t]{2mm}{\rotatebox[origin=c]{90}{NVD}}  &
    \includegraphics[align=c, width=0.08\textwidth, height = 37pt]{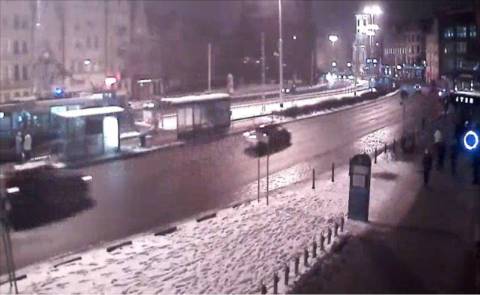} &
    \includegraphics[align=c, width=0.08\textwidth, height = 37pt]{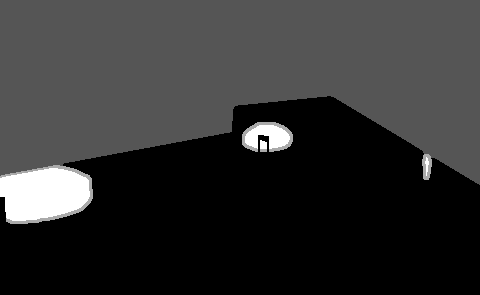} &
    \includegraphics[align=c, width=0.08\textwidth, height = 37pt]{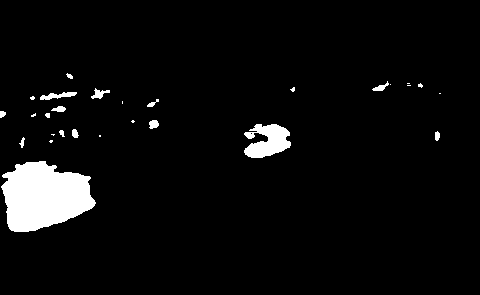} & 
    \includegraphics[align=c, width=0.08\textwidth, height = 37pt]{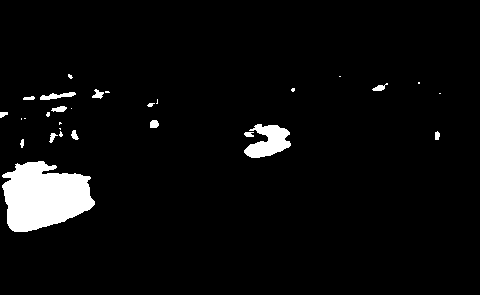} & 
    \includegraphics[align=c, width=0.08\textwidth, height = 37pt]{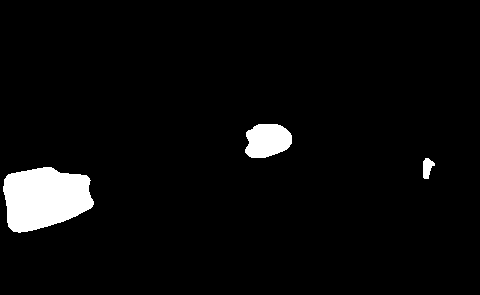} &
    \includegraphics[align=c, width=0.08\textwidth, height = 37pt]{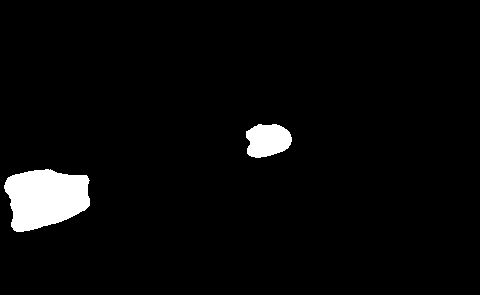} &
    \includegraphics[align=c, width=0.08\textwidth, height = 37pt]{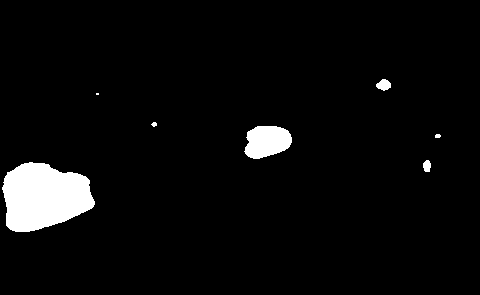} &
    \includegraphics[align=c, width=0.08\textwidth, height = 37pt]{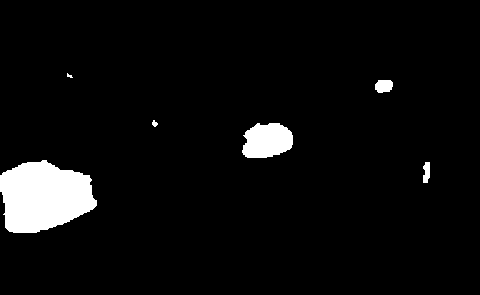} &
    \includegraphics[align=c, width=0.08\textwidth, height = 37pt]{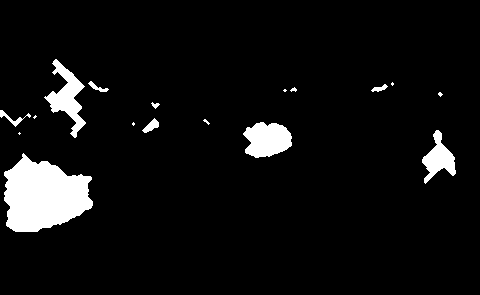} & 
    \Thickvrule{\includegraphics[align=c, width=0.08\textwidth, height = 37pt]{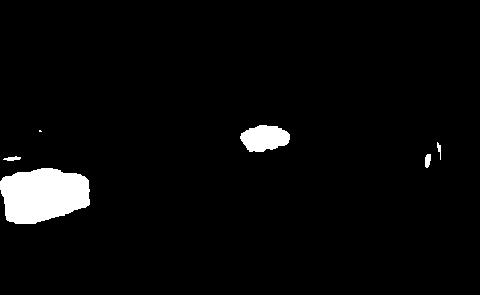}} &
    \includegraphics[align=c, width=0.08\textwidth, height = 37pt]{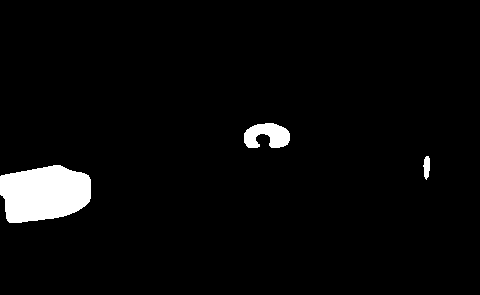} &
    \includegraphics[align=c, width=0.08\textwidth, height = 37pt]{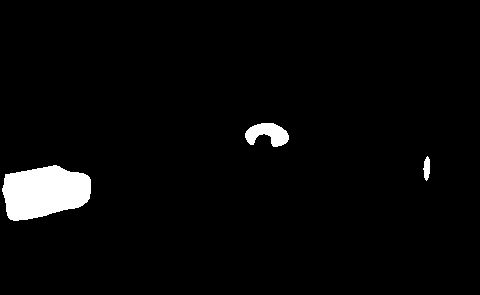} &
    \includegraphics[align=c, width=0.08\textwidth, height = 37pt]{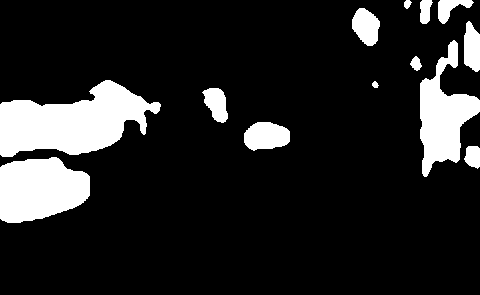} &
    \includegraphics[align=c, width=0.08\textwidth, height = 37pt]{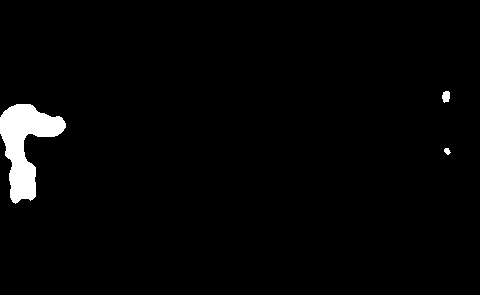} &
    \\
    
    \parbox[t]{2mm}{\rotatebox[origin=c]{90}{SHD}}  &
    \includegraphics[align=c, width=0.08\textwidth, height = 37pt]{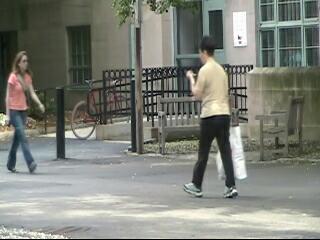} &
    \includegraphics[align=c, width=0.08\textwidth, height = 37pt]{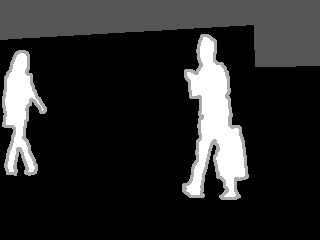} &
    \includegraphics[align=c, width=0.08\textwidth, height = 37pt]{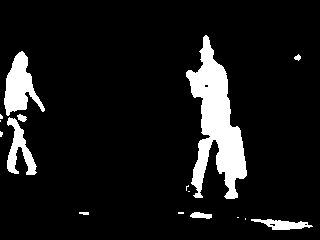} & 
    \includegraphics[align=c, width=0.08\textwidth, height = 37pt]{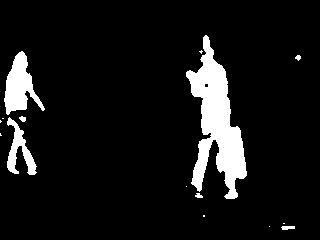} & 
    \includegraphics[align=c, width=0.08\textwidth, height = 37pt]{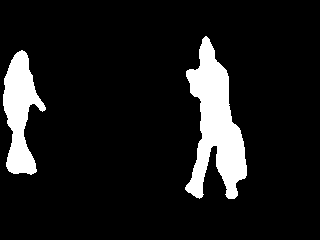} &
    \includegraphics[align=c, width=0.08\textwidth, height = 37pt]{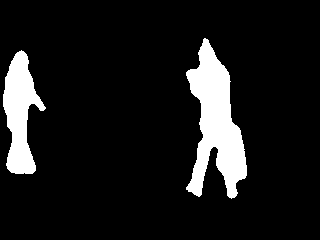} &
    \includegraphics[align=c, width=0.08\textwidth, height = 37pt]{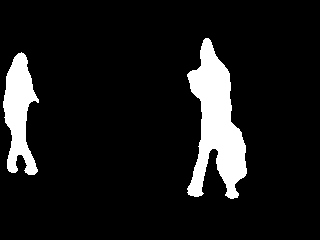} &
    \includegraphics[align=c, width=0.08\textwidth, height = 37pt]{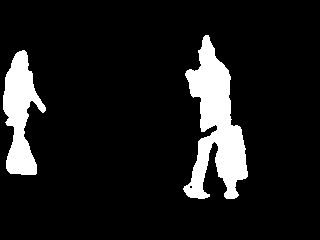} &
    \includegraphics[align=c, width=0.08\textwidth, height = 37pt]{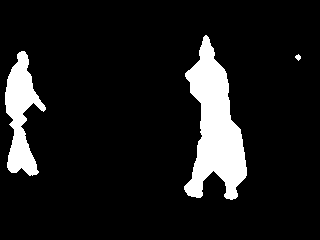} &     
    \Thickvrule{\includegraphics[align=c, width=0.08\textwidth, height = 37pt]{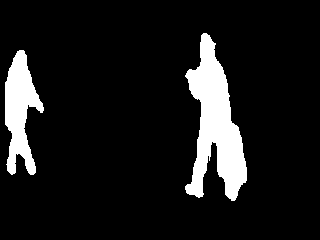}} &
    \includegraphics[align=c, width=0.08\textwidth, height = 37pt]{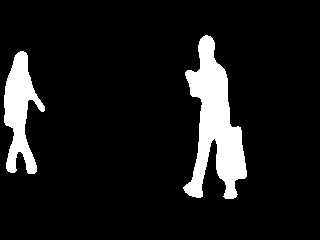} &
    \includegraphics[align=c, width=0.08\textwidth, height = 37pt]{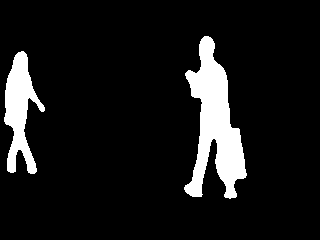} &
    \includegraphics[align=c, width=0.08\textwidth, height = 37pt]{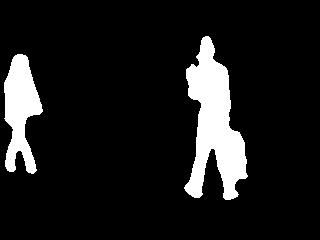} &
    \includegraphics[align=c, width=0.08\textwidth, height = 37pt]{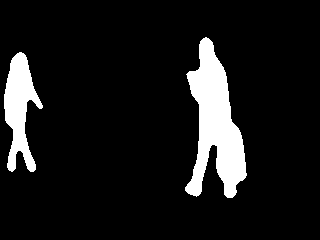} &
    \\
    
    \parbox[t]{2mm}{\rotatebox[origin=c]{90}{THM}}  &
    \includegraphics[align=c, width=0.08\textwidth, height = 37pt]{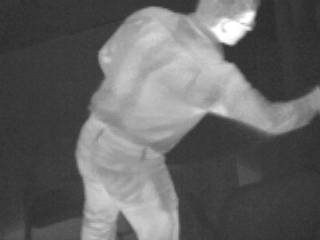} &
    \includegraphics[align=c, width=0.08\textwidth, height = 37pt]{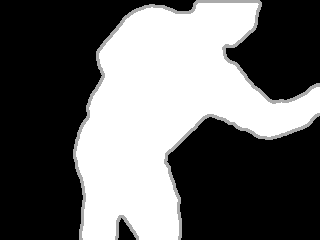} &
    \includegraphics[align=c, width=0.08\textwidth, height = 37pt]{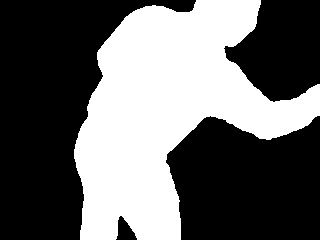} & 
    \includegraphics[align=c, width=0.08\textwidth, height = 37pt]{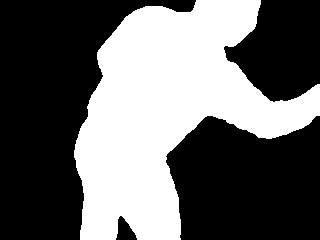} & 
    \includegraphics[align=c, width=0.08\textwidth, height = 37pt]{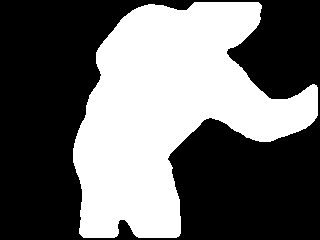} &
    \includegraphics[align=c, width=0.08\textwidth, height = 37pt]{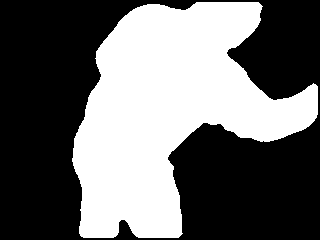} &
    \includegraphics[align=c, width=0.08\textwidth, height = 37pt]{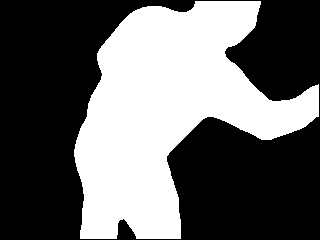} &
    \includegraphics[align=c, width=0.08\textwidth, height = 37pt]{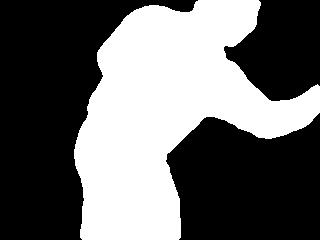} &
    \includegraphics[align=c, width=0.08\textwidth, height = 37pt]{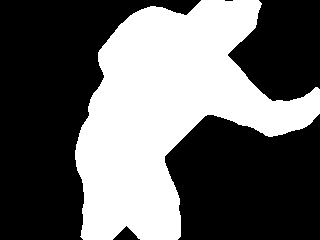} &     
    \Thickvrule{\includegraphics[align=c, width=0.08\textwidth, height = 37pt]{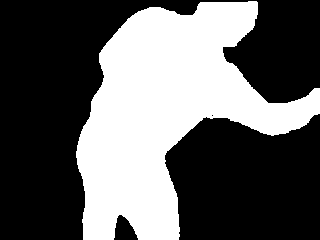}} &
    \includegraphics[align=c, width=0.08\textwidth, height = 37pt]{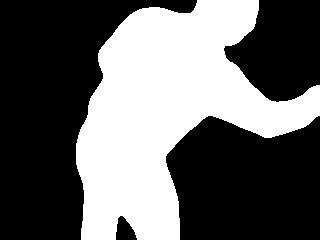} &
    \includegraphics[align=c, width=0.08\textwidth, height = 37pt]{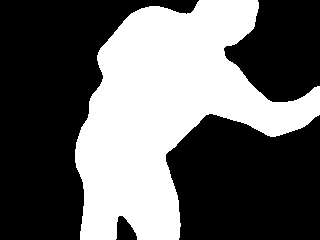} &
    \includegraphics[align=c, width=0.08\textwidth, height = 37pt]{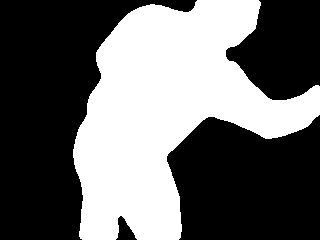} &
    \includegraphics[align=c, width=0.08\textwidth, height = 37pt]{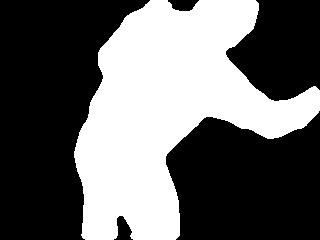} &
    \\
    
    \parbox[t]{2mm}{\rotatebox[origin=c]{90}{TBL}}  &
    \includegraphics[align=c, width=0.08\textwidth, height = 37pt]{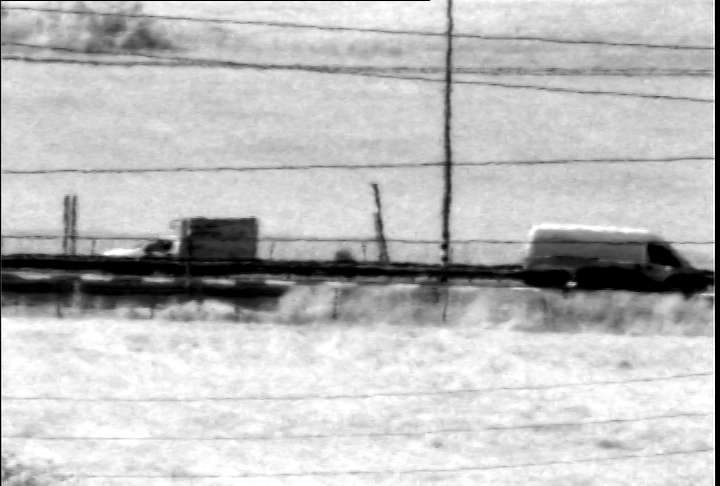} &
    \includegraphics[align=c, width=0.08\textwidth, height = 37pt]{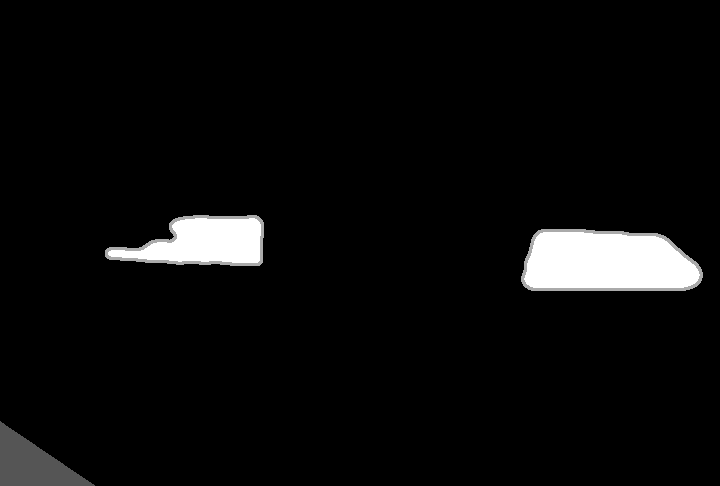} &
    \includegraphics[align=c, width=0.08\textwidth, height = 37pt]{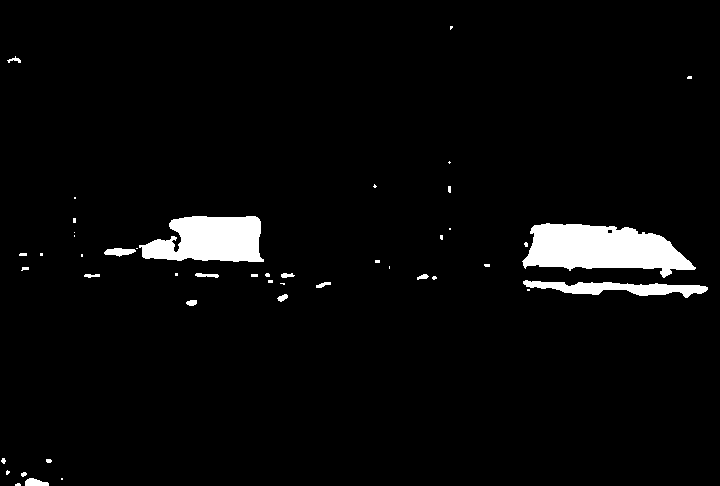} & 
    \includegraphics[align=c, width=0.08\textwidth, height = 37pt]{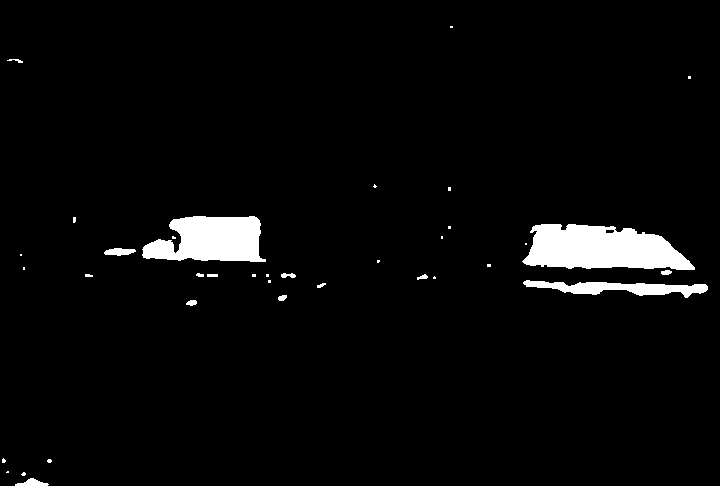} & 
    \includegraphics[align=c, width=0.08\textwidth, height = 37pt]{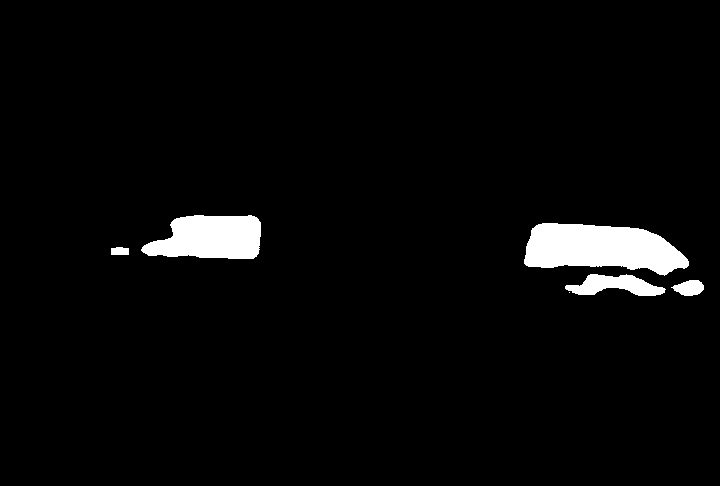} &
    \includegraphics[align=c, width=0.08\textwidth, height = 37pt]{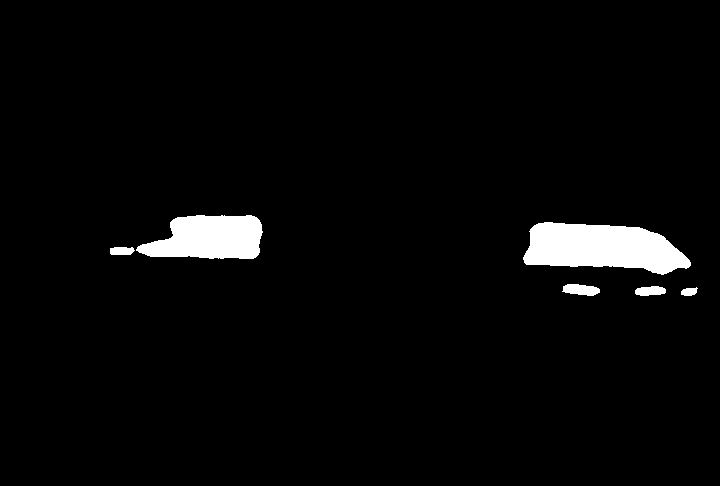} &
    \includegraphics[align=c, width=0.08\textwidth, height = 37pt]{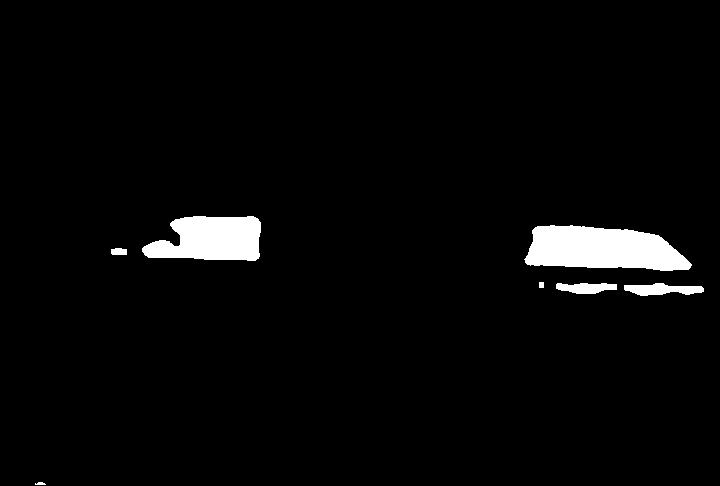} &
    \includegraphics[align=c, width=0.08\textwidth, height = 37pt]{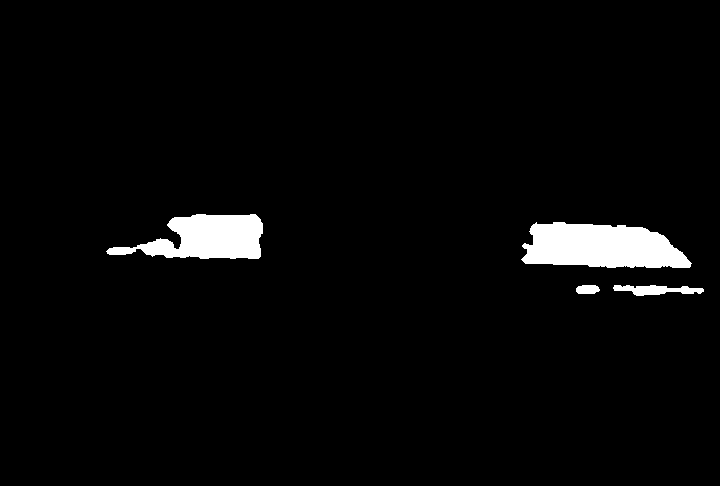} &
    \includegraphics[align=c, width=0.08\textwidth, height = 37pt]{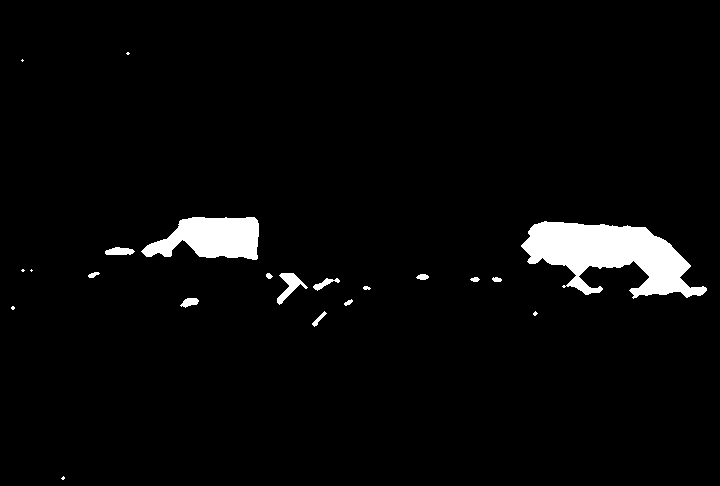} &     
    \Thickvrule{\includegraphics[align=c, width=0.08\textwidth, height = 37pt]{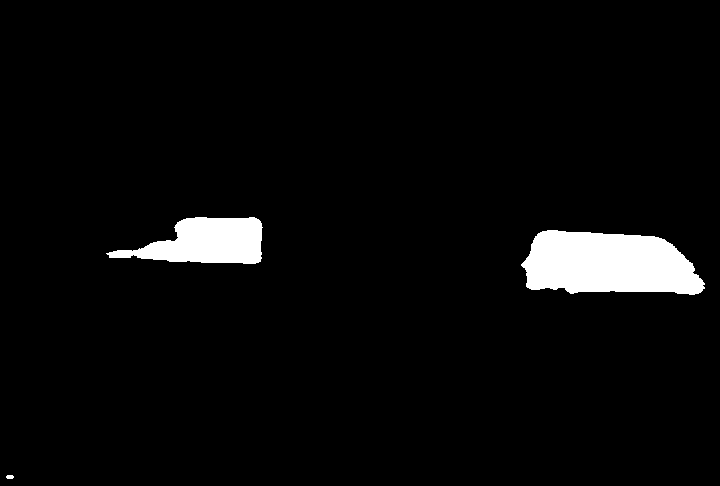}} &
    \includegraphics[align=c, width=0.08\textwidth, height = 37pt]{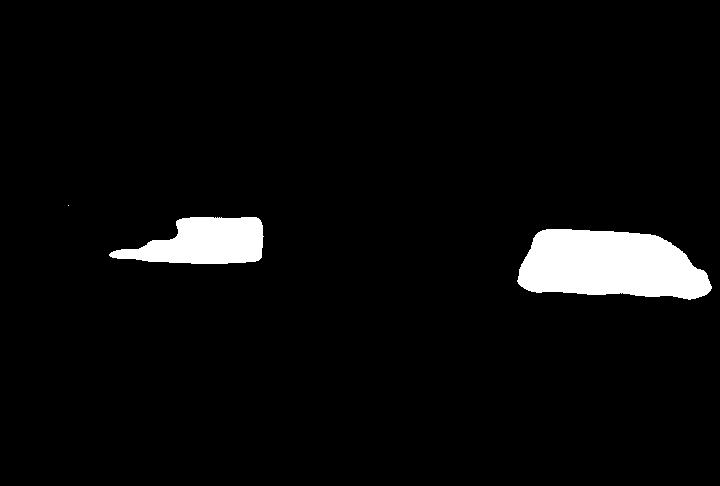} &
    \includegraphics[align=c, width=0.08\textwidth, height = 37pt]{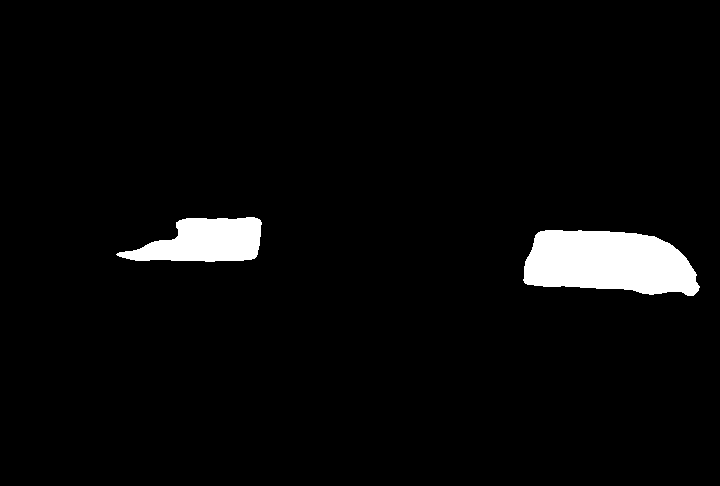} &
    \includegraphics[align=c, width=0.08\textwidth, height = 37pt]{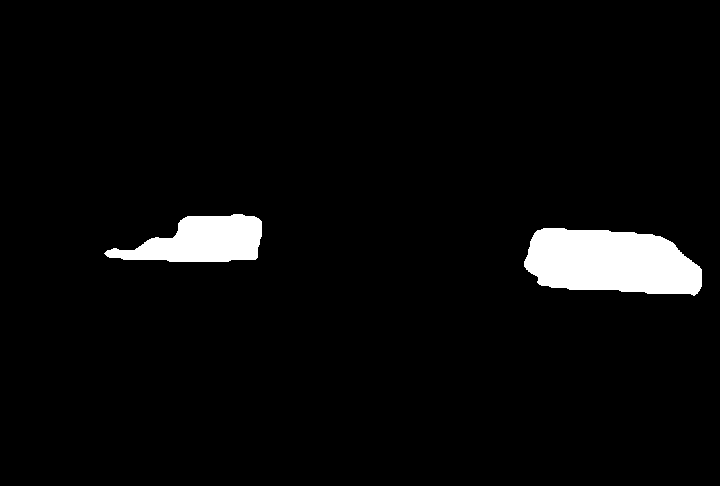} &
    \includegraphics[align=c, width=0.08\textwidth, height = 37pt]{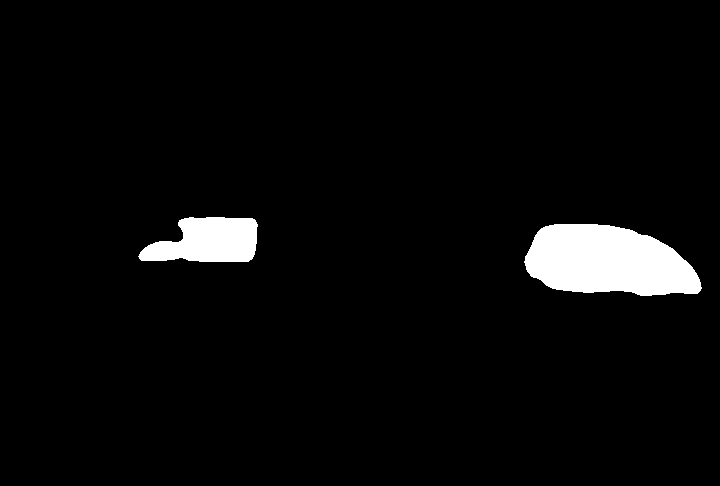} &
    \\[2ex]
    \arrayrulecolor{red}\cline{11-11}
    
    \end{tabular*}}
    \caption{Visual quality comparison for foreground detection on all video sequences in ten categories in CDnet 2014.
    The columns include: 
                   ($\star$) input frame, 
                   ($\diamond$) corresponding groundtruth foreground, 
                   (a) GMM -- S \& G,
                   (b) GMM -- Zivkovic,
                   (c) SuBSENSE,
                   (d) PAWCS,
                   (e) BMOG, 
                   (f) FTSG,
                   (g) SWCD,
                   (h) CDN-MEDAL-net, 
                   (i) FgSegNet\_S,
                   (j) FgSegNet\_v2
                   (k) Cascade CNN, 
                   (l) DeepBS.}
\label{fig:cdnet2014}
\vspace{-3mm}
\end{figure*}

The experimental results are summarized in Table \ref{tab:cdnet-fmeasure}, which highlights the F-measure quantitative results of our approach compared against several existing state-of-the-art approaches, along with Fig. \ref{fig:cdnet2014} that provides qualitative illustrations. For CDnet-2014, we pass over the \textit{PTZ} subdivision as this context deviates from popular application domains of video foreground extraction algorithms \cite{Garcia-Garcia2020}, which is not within our scope of consistent visual recordings from statically mounted cameras. In overall, despite its compact architecture, the proposed approach is shown to be capable of significantly outperforming unsupervised methods, and competing with complex deep-learning-based, supervised approaches in terms of accuracy. 


In comparison with unsupervised models built on the GMM background modeling framework like GMM -- Stauffer \& Grimson, GMM -- Zivkovic, BMOG and TensorMoG, the proposed approach is better augmented by the context-driven motion estimation plugin, without being constrained by simple thresholding schemes. Thus, it is able to provide superior F-measure results across the scenarios, especially on those where there are high degrees of noises on backgrounds like \textit{LFR}, \textit{NVD}, \textit{IOM}, \textit{CJT}, \textit{DBG} and \textit{TBL}. However, it is apparently a little worse than TensorMoG on \textit{BDW}, \textit{SHD}, \textit{IOM} and \textit{CJT}, which may be attributed to TensorMoG carefully tuned hyperparameters on segmenting foreground, thereby suggesting that the proposed method is still limited possibly by its architectural size and training data. 
Comparison of our proposed method is also conducted with other unsupervised methods that use mathematically rigorous approaches, such as SuBSENSE, PAWCS, FTSG, SWCD, to tackle scenarios commonly seen in real life (i.e. \textit{BSL}, \textit{DBG}, \textit{SHD}, and \textit{BDW}).
Nevertheless, F-measure results of the proposed approach around 0.90 suggests that it is able to outperform these complex unsupervised approaches, possibly ascribing to our use of hand-labeled data for explicitly enabling context capturing.

\begin{table}[t!]
\vspace{-3mm}
\caption{Result of quantitative evaluation on CDnet 2014 dataset}
\label{tab:quantitative-evaluation-cdnet-2014}
\centering
\resizebox{0.48\textwidth}{!}{
\begin{tabular}{l|llllll}
\toprule
& \multicolumn{1}{c}{\multirow{2}{*}{\textbf{Method}}}                         & \multicolumn{1}{c}{\textit{\textbf{Average}}} & \multicolumn{1}{c}{\textit{\textbf{Average}}} & \multicolumn{1}{c}{\textit{\textbf{Average}}} & \multicolumn{1}{c}{\textit{\textbf{Average}}} & \multicolumn{1}{c}{\textit{\textbf{Average}}} \\
&  & \multicolumn{1}{c}{\textit{\textbf{Recall}}} & \multicolumn{1}{c}{\textit{\textbf{FPR}}} & \multicolumn{1}{c}{\textit{\textbf{FNR}}} & \multicolumn{1}{c}{\textit{\textbf{PWC}}} & \multicolumn{1}{c}{\textit{\textbf{Precision}}} \\
\midrule
\parbox[t]{2mm}{\multirow{8}{*}{\rotatebox[origin=c]{90}{Unsupervised}}}
& GMM -- S \& G              & 0.6846                      & 0.0250                      & 0.3154                      & 3.7667                      & 0.6025                        \\
& GMM -- Zivkovic                         & 0.6604                      & 0.0275                      & 0.3396                      & 3.9953                      & 0.5973                        \\
& SuBSENSE                                & 0.8124                      & 0.0096                      & \textcolor{red}{$0.1876_{(1)}$}                      & 1.6780                      & 0.7509                        \\
& PAWCS                                   & \textcolor{blue}{$0.7718_{(3)}$}                      & \textcolor{red}{$0.0051_{(1)}$}                      & 0.2282                      & \textcolor{red}{$1.1992_{(1)}$}                      & \textcolor{green}{$0.7857_{(2)}$}                        \\
& TensorMoG                               & \textcolor{green}{$0.7772_{(2)}$}                      & 0.0107                      & \textcolor{blue}{$0.2228_{(3)}$}                      & 2.3315                      & \textcolor{red}{$0.8215_{(1)}$}                        \\
& BMOG                                    & 0.7265                      & 0.0187                      & 0.2735                      & 2.9757                      & 0.6981                        \\
& FTSG                                    & 0.7657                      & \textcolor{blue}{$0.0078_{(3)}$}                      & 0.2343                      & \textcolor{blue}{$1.3763_{(3)}$}                      & \textcolor{blue}{$0.7696_{(3)}$}                        \\
& SWCD                                    & \textcolor{red}{$0.7839_{(1)}$}                      & \textcolor{green}{$0.0070_{(2)}$}                      & \textcolor{green}{$0.2161_{(2)}$}                      & \textcolor{green}{$1.3414_{(2)}$}                      & 0.7527                        \\
\midrule
\parbox[t]{2mm}{\multirow{1}{*}{\rotatebox[origin=c]{90}{$\ast$}}}
& \textbf{CDN-MEDAL-net}             & \textbf{0.9232}                      & \textbf{0.0039                     } & \textbf{0.0768}                      & \textbf{0.5965}                      & \textbf{0.8724}                        \\ 
\midrule
\parbox[t]{2mm}{\multirow{6}{*}{\rotatebox[origin=c]{90}{Supervised}}}
 & FgSegNet\_S              & \textcolor{red}{$0.9896_{(1)}$} & \textcolor{green}{$0.0003_{(2)}$} & \textcolor{red}{$0.0104_{(1)}$} & \textcolor{green}{$0.0461_{(2)}$} & 0.9751 \\
 & FgSegNet                 & \textcolor{blue}{$0.9836_{(3)}$} & \textcolor{red}{$0.0002_{(1)}$} & \textcolor{blue}{$0.0164_{(3)}$} & \textcolor{blue}{$0.0559_{(3)}$} & 0.9758 \\
 & FgSegNet\_v2             & \textcolor{green}{$0.9891_{(2)}$} & \textcolor{red}{$0.0002_{(1)}$} & \textcolor{green}{$0.0109_{(2)}$} & \textcolor{red}{$0.0402_{(1)}$} & \textcolor{green}{$0.9823_{(2)}$} \\
 & Cascade CNN              & 0.9506 & 0.0032 & 0.0494 & 0.4052 & 0.8997 \\
 & DeepBS                   & 0.7545 & 0.0095 & 0.2455 & 1.9920  & 0.8332 \\
 & STAM                     & 0.9458 & \textcolor{blue}{$0.0005_{(3)}$} & 0.0542 & 0.2293 & \textcolor{red}{$0.9851_{(1)}$} \\
\bottomrule
\end{tabular}}{}
\resizebox{0.48\textwidth}{!}{\begin{tabular}{cccccc}      
\multicolumn{6}{p{10cm}}{\footnotesize ${}^{\ast}$Semi-Unsupervised; In each column, \textcolor{red}{$Red_{(1)}$} is for the best, \textcolor{green}{$Green_{(2)}$} {is for} the second best, and~\textcolor{blue}{$Blue_{(3)}$} {is for} the third best.}
\end{tabular}}{}
\vspace{-5mm}
\end{table}

In comparison with supervised approaches, the proposed approach is apparently competitive against the more computationally expensive state-of-the-arts. Our approach surpasses the generalistic methods of STAM and DeepBS on \textit{LFR} and \textit{NVD}, but it loses against both of these methods on \textit{SHD} and \textit{CMJ}, and especially is outperformed by STAM on many scenarios. While STAM and DeepBS are constructed using only 5\% of CDnet-2014, they demonstrate good generalization capability across multiple scenarios by capturing the holistic features of their training dataset. However, despite being trained on all scenarios, their behaviors showcase higher degrees of instability (e.g. with \textit{LFR}, \textit{NVD}) than our proposed approach on scenarios that deviate from common features of the dataset. Finally, as our proposed method is compared against similarly scene-specific approaches like FgSegNet’s and Cascade CNN, the results were within expectations for almost all scenarios that ours would not be significantly outperformed, as the compared models could accommodate various features of each sequence in their big architectures. However, our method surpasses even these computationally expensive to be at the top of the \textit{LFR} scenarios. This suggests that, with a background for facilitating motion segmentation as an input, our trained model can better tackle scenarios where objects are constantly changing and moving than even existing state-of-the-arts.

Table \ref{tab:quantitative-evaluation-cdnet-2014} presents evaluation metrics of a confusion matrix. Overall, with small training sets, CDN-MEDAL achieved decent results in Precision, Recall, FPR, FNR, PWC and a score of 0.8972 in average F-measure, which is much higher than any compared unsupervised approaches and can practically compete with other, more computationally expensive, supervised approaches despite its light-weighted structure.

\begin{table*}[!t]
\vspace{-5mm}
\setcounter{table}{4}
\caption{F - measure comparisons over the six sequences of Wallflower dataset with model parameters tuned on CDnet-2014}
\label{tab:wallflower-fmeasure}
\centering
\resizebox{0.85\textwidth}{!}{\begin{tabular}{l|l|llllll}
\toprule
& \multicolumn{1}{c|}{\textbf{Method}}                                          
& \textit{\textbf{Bootstrap}} & \textit{\textbf{LightSwitch}} & \textit{\textbf{WavingTrees}} & \textit{\textbf{Camouflage}}  & \textit{\textbf{ForegroundAperture}} & \textit{\textbf{TimeOfDay}} \\
\midrule
\parbox[t]{2mm}{\multirow{2}{*}{\rotatebox[origin=c]{90}{${}^{\mp}$UnS.}}}
& \multicolumn{1}{l|}{GMM -- Stauffer \& Grimson} & 0.5306    & 0.2296      & \textbf{0.9767}      & 0.8307      & 0.5778             & 0.7203    
\\ 
& SuBSENSE                                         & 0.4192    & 0.3201      & 0.9597      & 0.9535      & 0.6635             & 0.7107    
\\ 
\midrule
\parbox[t]{2mm}{\multirow{1}{*}{\rotatebox[origin=c]{90}{$\ast$}}}
& \multirow{1}{*}{\textbf{CDN-MEDAL-net}}                   & \textbf{0.7680}     & 0.5400        & 0.8156      & 0.9700        & \textbf{0.8401}             & \textbf{0.7429}    
\\
\midrule
\parbox[t]{2mm}{\multirow{2}{*}{\rotatebox[origin=c]{90}{${}^{\mp}$Sup.}}}
& DeepBS {[}33{]}                                  & 0.7479    & 0.6114      & 0.9546      & \textbf{0.9857}      & 0.6583             & 0.5494    
\\
& STAM                                             & 0.7414    & \textbf{0.9090}       & 0.5325      & 0.7369      & 0.8292             & 0.3429    
\\ 
\bottomrule
\end{tabular}}{}
\resizebox{0.85\textwidth}{!}{\begin{tabular}{cccccc}      
\multicolumn{6}{p{17cm}}{\footnotesize ${}^{\ast}$Semi-Unsupervised; ${}^{\mp}$UnS. = Unsupervised and Sup. = Supervised; In each column, \textbf{Bold} is for the best within each scenario.}
\end{tabular}}{}
\vspace{-3mm}
\end{table*}

\subsection{Results on Wallflower Benchmarks without Tuning}
Using the Wallflower dataset, we aim to empirically determine our proposed approach's effectiveness on unseen sequences, using only trained weights from scenarios of similar dynamics in CDnet-2014. The results apparently tend towards suggesting good degrees of our generalization from trained scenarios over to those unseen. Experimental evaluations are presented in Table \ref{tab:wallflower-fmeasure}, highlighting the F-measure quantitative results of our approach compared against some state-of-the-art methods in supervised, and unsupervised learning.

Specifically, on the \textit{Camouflage} scenario, our approach presents a very high score of 0.97 in terms of F-measure using the \textit{copyMachine} sequence of the \textit{SHD} scenario in CDnet-2014. As the model learns to distinguish between object motions and the shadow effects of \textit{copyMachine}, ours even extends to recognizing object motions of similar colors. Under Bootstrap where motions are present throughout the sequence, we employ the straight-forward background subtraction function learned via the clear features of static-view-versus-motion of \textit{highway} in \textit{BSL}, giving an F-score of 0.768. Likewise, the model’s capture of scene dynamics with \textit{office} of \textit{BSL}, \textit{backdoor} of \textit{SHD} and \textit{fountain02} of \textit{DBG} are extended towards respective views of similar features: \textit{ForegroundAperture} of clear motions against background, \textit{TimeOfDay} where there are gradual illumination changes and \textit{WavingTrees} of dynamic background motions, providing decently accurate results. On the other hand, the \textit{LightSwitch} scenario presents a big challenge where lightings are abruptly changed. As there is no scenario with this effect on the CDnet-2014 dataset, we chose the \textit{SHD} simply for its ability to distinguish objects clearly but the F-measure result is quite poor.

In comparison with existing methods whose aim are towards generalization like some unsupervised approaches GMM -- Stauffer \& Grimson, SuBSENSE, and CDnet-pretrained supervised approaches STAM, DeepBS, our proposed method yields very good results on \textit{Camouflage} and \textit{WavingTrees}, with even relatively better results on \textit{Bootstrap}, \textit{ForegroundAperture} and \textit{TimeOfDay}. While obviously this does not evidence that our approach is capable of completely better generalization from training than others, it does suggest that the proposed framework is able to excellently generalize to scenarios with dynamics similar to those learned, as supported by its relatively poor accuracy on \textit{LightSwitch}.

\subsection{Computational Speed Comparison}
\begin{figure}[!t]
    \vspace{-2mm}
    \centering
    \includegraphics[width=0.5\textwidth]{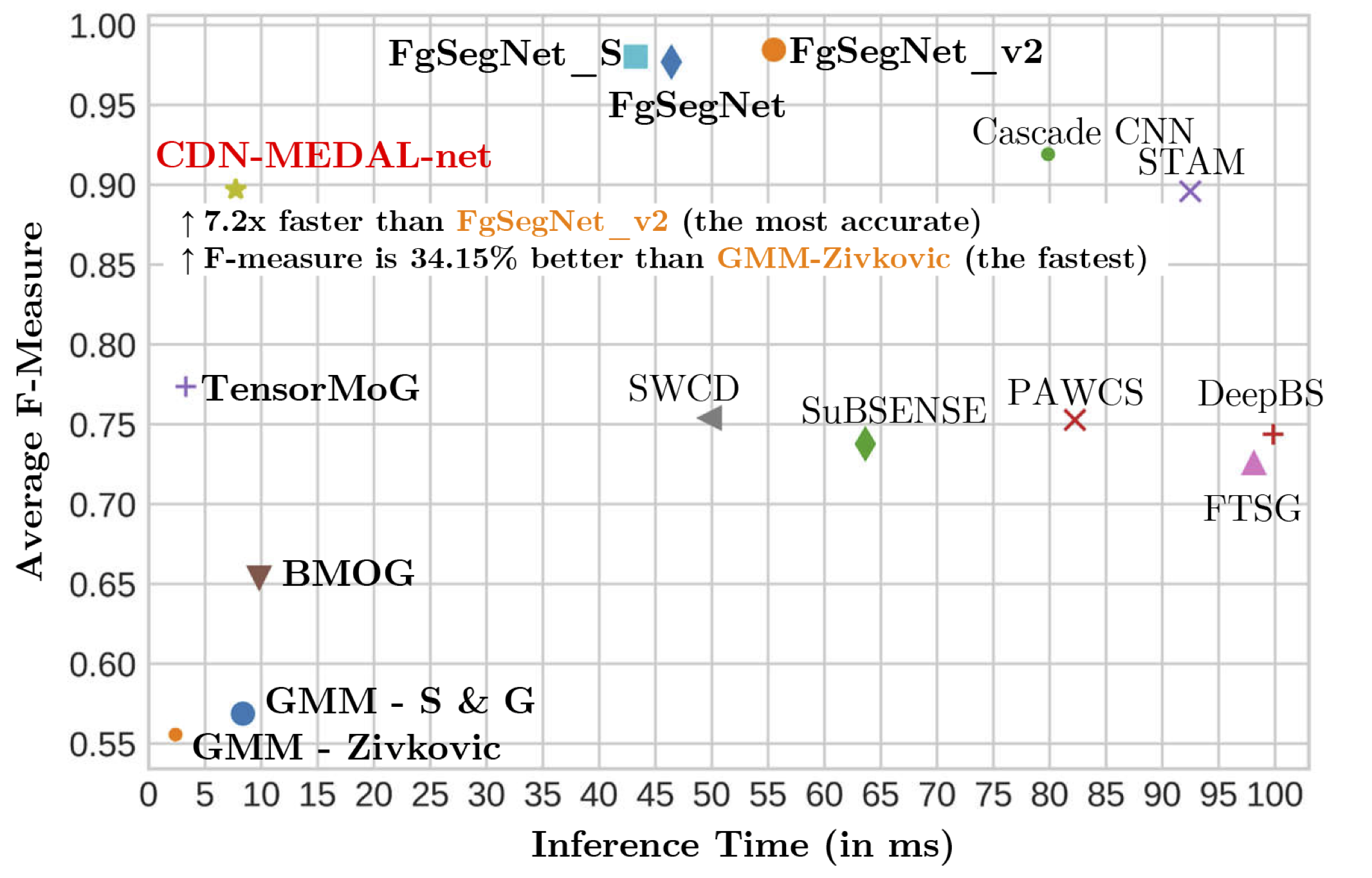}
    \caption{Computational speed and average F-measure comparison with state-of-the-art methods.}
    \label{figure:05}
    \vspace{-3mm}
\end{figure}

The proposed framework was implemented on a CUDA-capable machine with an NVIDIA GTX 1070 Ti GPU or similar, along with the methods that require CUDA runtime, i.e.,  TensorMoG, DeepBS, STAM, FgSegNet, and Cascade CNN. For unsupervised approaches, we conducted our speed tests on the configuration of an Intel Core i7 with 16 GB RAM. Our results are recorded quantitatively with execution performance in frame-per-seconds (FPS), and time (miliseconds) versus accuracy in Fig. \ref{figure:05}. At the overall speed of 129.4510 fps (from about 3,500 parameters), with CDN-GM (about 2,800 parameters) module processing at 402.1087 fps, CDN-MEDAL-net is much faster than other supervised deep learning approaches, of which the fastest - FgSegNet\_S - runs at 23.1275 fps. By concatenating estimations of background scenes with raw signals for foreground extraction, our approach makes such efficient use of hardware resources due of its completely lightweight architecture and the latent-space-limitation approach. In contrast, other DNNs architectures are burdened with a large number of trainable parameters to achieve accurate input-target mapping. Furthermore, the proposed scheme dominates the mathematically rigorous unsupervised methods frameworks in terms of speed and accuracy such as SuBSENSE, SWCD, and PAWCS, as their paradigms of sequential processing is penalized by significant penalties in execution. Significantly, the average speeds of the top three methods dramatically disparate. With the objective of parallelizing the traditional imperative outline of rough statistical learning on GMM, TensorMoG reformulates a tensor-based framework that surpasses our dual architecture at 302.5261 fps. On the other hand, GMM - Zivkovic's design focuses on optimizing its mixture components, thereby significantly trading off its accuracy to attain the highest performance. 

Notwithstanding, our proposed framework gives the most balanced trade-off (top-left-most) in addressing the speed-and-accuracy dilemma. Our model outperforms other approaches of top accuracy ranking when processing at exceptionally high speed, while obtaining good accuracy scores, at over 90\% on more than half of CDnet's categories and at least 84\%.

\section{Conclusion}
\label{section:conclusion}
This paper has proposed a novel, two-stage framework with a GMM-based CNN for background modeling, and a convolutional auto-encoder MEDAL-net to simulate input-background subtraction for foreground detection, thus being considered as a search space limitation approach to compress a model of DNNs, while keeping its high accuracy. Our first and second contributions in this paper include a pixel-wise, light-weighted, feed-forward CNN representing a multi-modular conditional probability density function of the temporal history of data, and a corresponding loss function for the CNN to learn from virtually inexhaustible datasets for approximating the mixture of Gaussian density function. In such a way, the proposed CDN-GM not only gains better capability of adaptation in contextual dynamics with humanly interpretable statistical learning for extension, but it is also designed in the tensor form to exploit modern parallelizing hardware. Secondly, we showed that incorporating such statistical features into MEDAL-net's motion-region extraction phase promises more efficient use of powerful hardware, with prominent speed performance and high accuracy, along a decent generalization ability using a small-scale set of training labels, in a deep non-linear scheme of only a few thousand latent parameters.

%


\bibliographystyle{IEEEtran}
\bibliography{references}

\end{document}